%% file: main.tex
\documentclass{article}

\PassOptionsToPackage{numbers}{natbib}

\usepackage[main, final]{neurips_2026}

\usepackage[utf8]{inputenc} %
\usepackage[T1]{fontenc}    %
\usepackage{url}            %
\usepackage{booktabs}       %
\usepackage{amsfonts}       %
\usepackage{nicefrac}       %
\usepackage{microtype}      %
\usepackage{xcolor}         %

\usepackage{graphicx}
\usepackage{booktabs}

\usepackage{color,colortbl}
\definecolor{cvprblue}{rgb}{0.21,0.49,0.74}
\usepackage[pagebackref,breaklinks,colorlinks,citecolor=cvprblue]{hyperref}

\input{_preamble}

\input{_constants}

\usepackage{ragged2e}
\begin{document}

\maketitle

\input{figures/00_teaser/figure}

\input{sections/00_abstract}

\input{sections/01_intro}

\input{sections/02_related}

\input{sections/03_method}
\input{sections/04_result}

\input{sections/05_discussion}

\section*{Acknowledgments}
This work was supported by a research grant from the Youlchon Foundation, Institute of Engineering Research at Seoul National University, the National AI Project of the Ministry of Science and ICT and the AIIS.,NRF grant funded by the Korean government (MSIT) [No. RS-2026-25540546], and IITP grant funded by the Korea government (MSIT) [No. RS-2024-00439854, No. RS-2025-25442338, No. RS-2025-25441838, No. 2022-0-00156, and No. RS-2021-II211343]. H. Joo is the corresponding author.

{
    \small
    \bibliographystyle{ieeenat_fullname}
    \bibliography{main}
}

\input{sections/10_supple.tex}

\end{document}

%% file: _preamble.tex
\usepackage{bm}
\usepackage{xspace}
\usepackage{amsmath}
\usepackage[capitalize]{cleveref}
\usepackage{caption}

\usepackage{algorithm}
\usepackage{algpseudocode}

\usepackage{amsfonts}
\usepackage{amssymb}
\usepackage{enumitem}
\usepackage{etoc}

\algnewcommand\algorithmicinput{\textbf{Input:}}
\algnewcommand\Input{\item[\algorithmicinput]}
\algnewcommand\algorithmicoutput{\textbf{Output:}}
\algnewcommand\Output{\item[\algorithmicoutput]}

\renewcommand{\vec}[1]{\bm{#1}}

\definecolor{tabfirst}{rgb}{1, 0.7, 0.7} %
\definecolor{tabsecond}{rgb}{1, 0.85, 0.7} %
\definecolor{tabthird}{rgb}{1, 1, 0.7} %

\newcommand{\figref}[1]{Fig.~\ref{#1}}
\newcommand{\tabref}[1]{Tab.~\ref{#1}}
\newcommand{\eqnref}[1]{Eq.~(\ref{#1})}
\newcommand{\secref}[1]{Sec.~\ref{#1}}

%% file: _constants.tex
\def\modelname{SimuScene}
\def\dataname{GenWild} %
\def\paperTitle{Simulation-Ready Compositional \\3D Scene Reconstruction from a Single Image}
\title{\paperTitle}

\author{
    \textbf{Inhee Lee}$^{*}$\quad
    \textbf{Sangwon Baik}$^{*}$\quad
    \textbf{Sungjoo Kim}\\[-0.1em]
    \textbf{Hyeonwoo Kim}\quad
    \textbf{Hyunsoo Cha}\quad
    \textbf{Hanbyul Joo}$^{\dagger}$\\
    [0.4em]
    \normalfont Seoul National University\\
    [0.3em]
    \normalfont $^*$Equal Contribution \quad $\dagger$Corresponding Author\\
    [0.3em]
    \normalfont{\ttfamily\small \{ininin0516,bsw1907,masterninja,hwkim408,243stephen,hbjoo\}@snu.ac.kr}\\
    [0.3em]
    \normalfont{\ttfamily\small \href{https://github.com/snuvclab/SimuScene}{\color{magenta}{https://github.com/snuvclab/SimuScene}}}
}

%% file: figures/00_teaser/figure.tex
\begin{figure}[h]
    \vspace{-20pt}
    \centering
    \includegraphics[width=\textwidth]{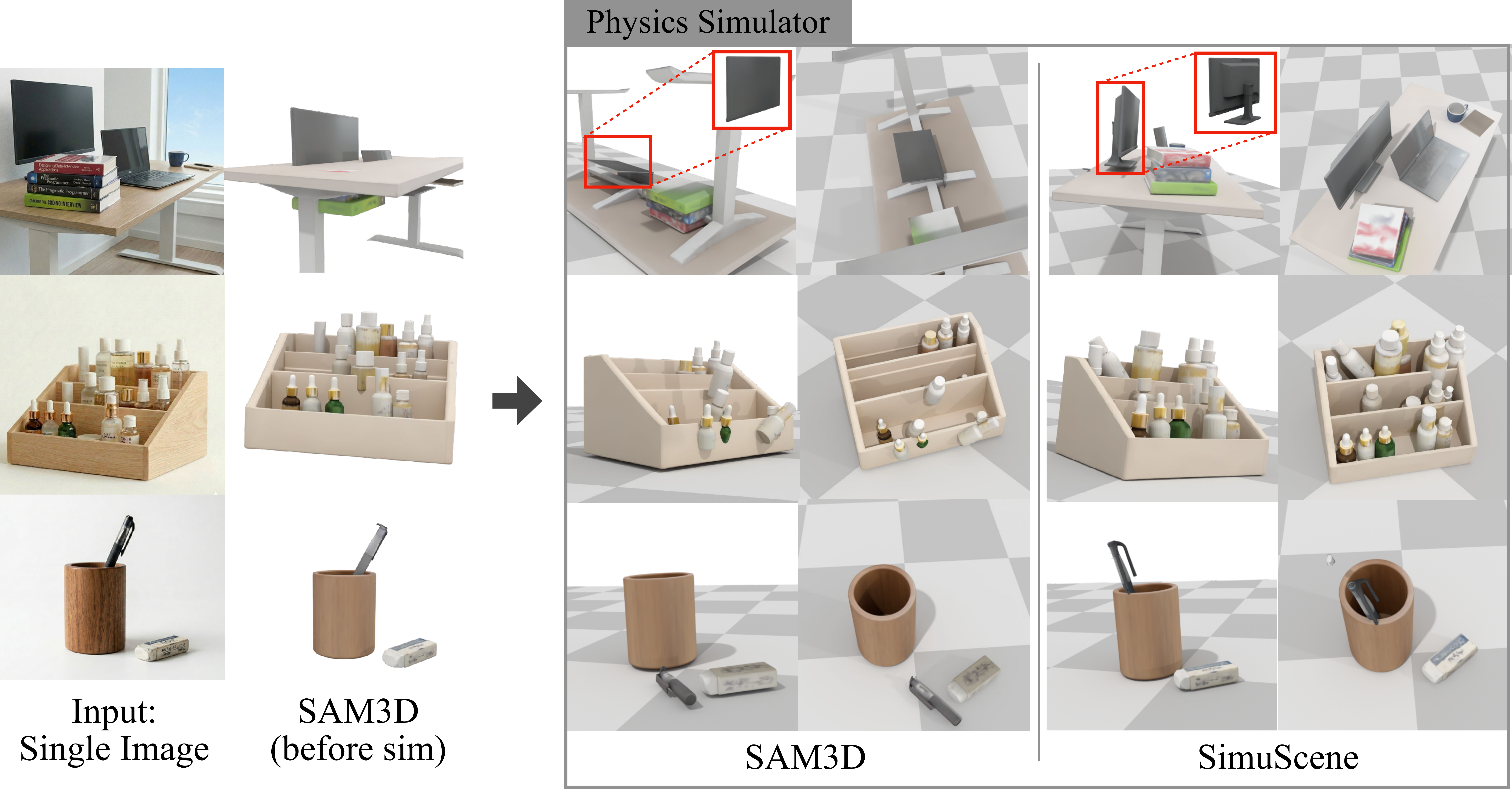}
    \caption{\textbf{\modelname} reconstructs simulation-ready compositional 3D scenes from a single image by using physics simulation to correct object shape and pose. Compared with conventional 3D lifters such as SAM3D~\cite{sam3d}, our method handles heavy occlusion and produces stable layouts. Red boxes highlight physics-guided shape corrections.
    }
    \label{fig:teaser}
\end{figure}

%% file: sections/00_abstract.tex
\begin{abstract}
    Reconstructing interactive, simulation-ready 3D scenes from a single image is a critical bottleneck for robotic manipulation. While recent single-image lifters recover plausible per-object shapes, composing them yields scenes that collapse under physical simulation due to interpenetrating, hovering, or sinking objects. Existing physics-aware compositional reconstruction methods mainly address this as a post-hoc layout correction, leaving the underlying geometric errors unresolved. To address this, we introduce SimuScene, a compositional 3D reconstruction pipeline that puts physics in the loop of shape and layout estimation. Rather than using physics merely for layout cleanup, we utilize the physics engine as a diagnostic measurement tool during the generative process itself. By diagnostically simulating reconstructed objects under gravity, we convert penetration and support failures into quantitative correction signals that drive amodal shape resampling, with gravity-axis stretching as a lightweight correction for minor errors. This physics-informed feedback loop mitigates accumulated reconstruction errors and produces a stable, simulation-ready compositional 3D scene. Extensive experiments demonstrate state-of-the-art performance on physical stability and geometric alignment benchmarks. We further highlight SimuScene's utility by deploying reconstructed environments in humanoid control and robot-arm manipulation tasks.
\end{abstract}

%% file: sections/01_intro.tex
\section{Introduction}

A single photograph of a cluttered desk contains, in principle, everything a robot needs to practice on it: the table to push against, the cup to grasp, the book to slide aside. In practice, however, current methods fall short of extracting this. Single-image scene methods either generate the entire view holistically, as novel views or a single inseparable 3D representation~\cite{gen3c, seva, viewcrafter, gao2024cat3d, wonderworld, luciddreamer}, or lift each object independently and produce configurations that collapse the moment a physics simulator releases them~\cite{midi,scenegen,sam3d}. Objects sink through tables, hover above shelves, or wedge into one another with overlapping geometry. The difficulty is twofold. A single image is fundamentally ambiguous about occluded geometry and absolute depth, and the physical plausibility that would resolve this ambiguity is not a property any existing dataset readily teaches. 

Our goal is to reconstruct a physically plausible, simulation-ready compositional 3D scene from a single image, an output that drops directly into a physics simulator and supports downstream tasks such as reinforcement and imitation learning for robot manipulation, or a VR/AR setup as an interactive 3D scene that responds correctly to contact and gravity. 
Recent single-image approaches~\cite{sam3d} make substantial progress by lifting each detected object into a complete 3D shape. 
While these results are visually convincing, they are often not physically plausible: once placed in a simulator, objects fall through their supports or explode out of contact, meshes interpenetrate, support relationships break, and objects sink or hover, due to errors in both occlusion-induced shape completion and monocular object pose estimation. Recent efforts~\cite{cast} begin to incorporate physics into the reconstruction process, yet they apply it only at the layout level, adjusting where each object sits while leaving the underlying shapes untouched. When the geometry itself is incorrect, post-hoc layout adjustment cannot fully recover a plausible configuration. This failure mode also exposes an opportunity: the physics simulation that reveals these artifacts also measures them. For example, the distance an object falls, or the depth at which it interpenetrates a neighbor, gives an important clue to the geometric error that a single view alone cannot supply. Rather than treating these signals as artifacts to clean up at the end, we use them as diagnostic measurements that drive shape correction during reconstruction itself.

We realize this idea in \modelname, a single-image compositional 3D scene reconstruction pipeline that puts \emph{physics in the loop} of shape and layout estimation. In our pipeline, the physics simulation acts as a diagnostic loop that drives shape and scale correction, not a post-hoc cleanup. The gravity-direction displacement at first contact exposes whether the lifted shape is too short, too tall, or grossly mis-shaped, and feeds directly into the geometry update. We feed this simulator feedback into a resampling stage that recovers more plausible 3D shapes even under heavy occlusion. We further treat the simulator as a source of evidence for shape itself, using the simulated configuration to define the target extent of hidden geometry and thereby reduce the residual uncertainty that occlusion leaves behind. Building on these two roles of physics, \modelname{} composes a scene through an iterative, sequential \emph{physics in the loop} procedure that reconstructs one object at a time. Once loaded into a physics simulator, our reconstructions achieve state-of-the-art performance on physical-plausibility and reconstruction-quality metrics. We further demonstrate the resulting object-complete scenes driving downstream applications including humanoid control and robot-arm manipulation tasks.

\noindent
We make following contributions:
(1)~\textbf{Physics-in-the-loop diagnostic simulation.} A sequential, per-object protocol integrates physical dynamics directly into reconstruction, converting violations (e.g., interpenetration, gravity-induced displacement) into actionable diagnostic signals.
(2)~\textbf{Physics-informed shape correction.} We use simulated displacement to select between two geometry updates. OBB-guided amodal resampling improves shape quality for large displacements, while gravity-axis stretching reduces computation for small ones.
(3)~\textbf{Extensive experimentation.} 
We provide comprehensive evaluations across diverse datasets and metrics, and show that our reconstructed simulation-ready 3D scenes support downstream robotics applications (i.e. humanoid control policy learning, robot-arm manipulation)

%% file: sections/02_related.tex
\section{Related Work}
\label{sec:related_work}
\paragraph{Multi-Object 3D Reconstruction from Cluttered Images.}
Recovering a \emph{structured} 3D scene from a single RGB image is ill-posed under occlusion, depth ambiguity, and inter-object interactions, originally tackled by analysis-by-synthesis search over CAD exemplars under structural and physical constraints~\cite{huang2018holistic}.
End-to-end learning now produces coherent multi-object geometry from a single view via feed-forward regression with non-intersecting outputs~\cite{popov2020corenet}, multi-instance diffusion with cross-instance attention~\cite{midi}, hierarchical isometric-view amodal completion~\cite{dong2025hiscene}, or promptable scene-level reconstruction~\cite{sam3d}, while divide-and-conquer pipelines reconstruct each object independently and reassemble them via iterative occlusion removal~\cite{aguina2026seeing}, differentiable optimal-transport alignment~\cite{han2025reparo}, or feed-forward decoupling of appearance, rotation, scale, and translation~\cite{hu2025flash}.
3D-Fixer~\cite{3D-Fixer} further addresses occluded geometry through coarse-to-fine in-place completion, using observed partial geometry as a spatial anchor without explicit pose alignment.
Other related systems decompose scenes from multi-view captures~\cite{ni2025decompositionalneuralscenereconstruction,yang2025instascenecomplete3dinstance}, retrieve similar assets instead of reconstructing input-faithful geometry~\cite{dai2024automated}, or rebuild scenes as graphics programs through agentic multimodal reasoning~\cite{yin2026visionasinversegraphicsagentinterleavedmultimodal}. These systems differ from our single-image setting in input modality or reconstruction target.
Despite improving geometric fidelity and instance separation, the resulting assets routinely interpenetrate, float, or fail to satisfy support relations, leaving the scenes unsuitable for downstream simulation and embodied use.

\paragraph{Physics-Aware Scene Reconstruction and Simulation-in-the-Loop Refinement.}

Physics constraints have long shaped reconstruction, from early scene parsers that encode support and collision feasibility in the inference objective~\cite{huang2018holistic} to recent physics-aware and differentiable-simulation pipelines that jointly optimize geometry, pose, material, or appearance via physically plausible implicit surfaces~\cite{ni2024phyrecon}, MPM-driven compositional Gaussian generation~\cite{yan2024phycage}, amodal MPM-coupled scene reconstruction~\cite{chen2025physgen3d}, or human-scene depth-alignment and contact priors~\cite{physic}, with simulator feedback also serving as a reward-style alignment signal for image-to-3D generators~\cite{dso}.
At the scene level, relation-graph rigid-body solvers~\cite{cast} and simulator-in-the-loop assembly~\cite{xia2026simreconsimreadycompositionalscene} resolve penetration and floating artifacts in assembled layouts, while text-conditioned generators embed SDF collision avoidance, gravity, or differentiable rigid-body simulation directly in the generation loop to produce intersection-free, statically stable scenes~\cite{scenethesis,phipg,pat3d}.

\paragraph{Direct Preference Optimization for Generative Models.}

Direct Preference Optimization (DPO) replaces the reward modeling and on-policy RL step of RLHF~\cite{rlhf-instructgpt} with a simple preference-classification objective derived from a closed-form reparameterization of the Bradley--Terry reward model~\cite{dpo}. 
Subsequent language-model alignment methods generalize this objective~\cite{ipo}, replace pairwise preferences with binary utility feedback~\cite{kto}, integrate preference optimization into supervised fine-tuning~\cite{orpo}, or remove the reference model through a length-normalized reward with a target margin~\cite{simpo}. 
For visual generation, Diffusion-DPO~\cite{diffusion-dpo} adapts DPO to diffusion models by replacing autoregressive log-likelihood ratios with denoising-loss differences, while related methods formulate denoising as a policy optimization problem~\cite{d3po,ddpo} or introduce step-wise preference comparisons~\cite{spo}. 
At the 3D level, DSO~\cite{dso} uses rigid-body simulation feedback to fine-tune image-to-3D generators for physical soundness with DPO or DRO objectives. 
Inspired by DSO and Diffusion-DPO, we fine-tune the SAM3D shape branch with a flow-matching DPO objective over synthetic occlusion-versus-amodal-completion pairs.
Unlike DSO, which optimizes standalone object stability, our preference signal targets occlusion-robust shape resampling for objects in cluttered scenes.

%% file: sections/03_method.tex
\input{figures/30_pipeline/figure}

\section{Method}\label{sec:method}

Given a single RGB image $\vec{I}$, our goal is to reconstruct a simulation-ready compositional 3D scene $\vec{\mathcal{S}}$, visually aligned with $\vec{I}$ and physically consistent under gravity:
\begin{equation}\label{eq:scene_def}
  \vec{\mathcal{S}} = \{\vec{\mathcal{M}}_i, s_i, \mathbf{t}_i, \mathbf{R}_i\}_{i=1}^N, 
\end{equation}
where $\vec{\mathcal{M}}_i$ is a 3D canonical mesh for $i$-th object, $s_i \in \mathbb{R}^+$ is an isotropic scale of $\vec{\mathcal{M}}_i$, $\mathbf{t}_i \in \mathbb{R}^3$ and $\mathbf{R}_i \in \mathrm{SO}(3)$ are a translation vector and a rotation matrix of $\vec{\mathcal{M}}_i$, and $N$ is the number of objects in the scene.
Monocular ambiguity and occlusion often cause penetration, floating, and toppling. We therefore use diagnostic physics signals to refine object poses and correct geometry through Oriented Bounding Box (OBB)-guided stretching or resampling.
\figref{fig:pipeline} summarizes our pipeline.

\subsection{Decomposed Scene Initialization}\label{sec:base_only}
Because base structures (e.g., tables, shelves) support most objects in cluttered scenes, we reconstruct them separately from other objects.
We first remove non-base objects with an MLLM-based image editor~\cite{gemini} and run SAM3D~\cite{sam3d} on the decluttered image $\vec{I}_{\mathrm{base}}$ to obtain fixed base colliders.
For the remaining objects, RAM++~\cite{rampp} and a VLM~\cite{gpt4} generate instance labels, SAM3~\cite{sam3} produces masks $\vec{M}_i$, and SAM3D, denoted by $\Phi$, lifts each instance:
\begin{equation}
  \Phi(\vec{I}, \vec{M}_i, \vec{I}_{\vec{M}_i}, \vec{D})
  =
  (\vec{\mathcal{M}}_i^\mathrm{init}, s_i^\mathrm{init}, \mathbf{t}_i^\mathrm{init}, \mathbf{R}_i^\mathrm{init}),
\end{equation}
where $\vec{M}_i$ is the segmentation mask of $i$-th object, $\vec{I}_{\vec{M}_i}$ is the masked image crop, and $\vec{D}$ is single-view depth from MoGe~\cite{wang2025moge}. $\vec{\mathcal{M}}_i^\mathrm{init}$ is in SAM3D's canonical frame, with its AABB centered at the origin and the longest side normalized to $1$.
Beyond geometry, we extract semantic information that proves critical for physics. The VLM tags each object with one of three pose-DoF labels, \textit{free} (6-DoF), \textit{point-anchored} (3-DoF rotation about a wall point), or \textit{line-anchored} (1-DoF rotation about the anchor line). These tags tell the simulator which constraints to apply to each object.
Since the floor and walls are fixed after initial estimation, we omit them from \eqnref{eq:scene_def} for simplicity. See \secref{sec:supp_pp} for additional details on scene decomposition.

\subsection{Pose Refinement Before Simulation}\label{sec:refinement}
SAM3D produces plausible shapes but in its own canonical frame, with poses only approximately consistent with the image. Handing these poses directly to a physics simulator amplifies even small rotational errors into severe penetration and contact artifacts.
We therefore refine each object's pose against image evidence before simulation.
We obtain depth $\vec{D}$ and camera intrinsics $\vec{K}$ from $\vec{I}$ using MoGe~\cite{wang2025moge}, and back-project the pixels inside the segmentation mask $\vec{M}_i$ to form the object point cloud $\vec{\mathcal{P}}_i \in \mathbb{R}^{P \times 3}$. We first solve for $(s_i^*, \mathbf{t}_i^*)$ by aligning the canonical mesh $\vec{\mathcal{M}}_i^\mathrm{init}$ with $\vec{\mathcal{P}}_i$ (see \secref{sec:supp_pp} for details).
Rotation, however, cannot be reliably recovered from point-cloud alignment alone, so we hand the result to FoundationPose~\cite{wen2024foundationpose} for refinement:
\begin{equation}\label{eq:fdp_refine}
  (\mathbf{t}^{\mathrm{pre}}_i, \mathbf{R}^{\mathrm{pre}}_i)
  =
  \mathrm{FoundationPose}
  \bigl(\vec{I}, \vec{D}, \vec{K}, \vec{M}_i, s_i^*\vec{\mathcal{M}}^{\mathrm{init}}_i, \mathbf{t}_i^*, \mathbf{R}^{\mathrm{init}}_i \bigr),
\end{equation}
where $s_i^*\vec{\mathcal{M}}^{\mathrm{init}}_i$ denotes $\vec{\mathcal{M}}^{\mathrm{init}}_i$ scaled by $s_i^*$, and $\mathbf{t}_i^*$ and $\mathbf{R}^{\mathrm{init}}_i$ from the previous alignment are used for the initials. 
The pre-simulation scene is then composed as:
\begin{equation}\label{eq:pre_sim_scene}
  \vec{\mathcal{S}}^\mathrm{pre} = \{\vec{\mathcal{M}}_i^{\mathrm{init}}, s_i^*, \mathbf{t}_i^{\mathrm{pre}}, \mathbf{R}_i^{\mathrm{pre}}\}_{i=1}^N.
\end{equation}

\subsection{Diagnostic Simulation: Physics as a Probe}\label{sec:physics}

The pre-simulation scene $\vec{\mathcal{S}}^\mathrm{pre}$ is the image-faithful configuration we can produce from perception alone, but it is not yet physically valid. It still contains penetrations and floating objects rooted in monocular reconstruction ambiguity, and as shown in \figref{fig:ablation}, naively dropping everything under gravity produces chaotic dynamics in which mutually penetrating objects push each other into impossible configurations. 
We use physics simulation to diagnose these residual errors: we first resolve inter-object penetration by displacing the target object along a simulator-informed direction. Then, after releasing the object under gravity, the resulting displacement and contact signals expose unresolved gravity-axis shape and support errors that drive the shape-correction stage (\secref{sec:shape_correction}).

To keep the diagnostic signal clean and avoid joint-settling deadlocks, we compose the scene with a sequential protocol. Objects are processed in ascending order of the lowest corner of their world-frame bounding box along the gravity direction. For each active object $i$, only that object is dynamic, while previously placed objects are frozen as static colliders, and the simulation halts at the first frame of contact. This prevents mutually penetrating objects from pushing each other into impossible configurations and avoids unstable trajectories that drift far from the image-aligned pose.

\noindent\textbf{Penetration Resolution.}
For simplicity, we assume that objects are indexed according to the sequential processing order, so that object $i$ is processed after objects $1,\ldots,i-1$.
For each active object $i$, the simulator generates a penetration-resolving displacement $\Delta \mathbf{t}^{\mathrm{pen}}_{i}$ to prevent overlap with the previously processed objects $1,\ldots,i-1$.
For the first object, the procedure resolves penetration with the floor and any walls.
The scene after penetration resolution, denoted by $\vec{\mathcal{S}}^{\mathrm{pen}}$, is formulated as
\begin{equation}\label{eq:post_pene_scene}
  \vec{\mathcal{S}}^\mathrm{pen}
  =
  \{\vec{\mathcal{M}}_i^{\mathrm{init}}, s_i^*, \mathbf{t}_i^{\mathrm{pre}} + \Delta \mathbf{t}^{\mathrm{pen}}_{i}, \mathbf{R}_i^{\mathrm{pre}}\}_{i=1}^N.
\end{equation}
Objects whose pose-DoF tag is \textit{point-anchored} or \textit{line-anchored} are anchored at their current positions using the procedure described in \secref{sec:supp_sim}.
Accordingly, as discussed in \secref{sec:base_only}, objects tagged as \textit{point-anchored} are restricted to 3-DoF rotation about the anchor point, while objects tagged as \textit{line-anchored} are restricted to 1-DoF rotation about the anchor line.

\noindent\textbf{Gravity-Based Diagnostics.}
We then sequentially release each active object $i$ under gravity.
If the pose-DoF tag of the object is \textit{free}, the object is frozen at the first simulation frame in which it contacts any surface, yielding the gravity-induced displacement $\Delta \mathbf{t}^{\mathrm{grv}}_{i}$.
For wall-anchored objects, i.e., those tagged as \textit{point-anchored} or \textit{line-anchored}, the object is simulated under its corresponding rotational constraint until it settles under gravity, yielding the gravity-induced rotation $\Delta \mathbf{R}^{\mathrm{grv}}_{i}$.
The resulting post-simulation scene $\vec{\mathcal{S}}^{\mathrm{sim}}$ is formulated as
\begin{equation}\label{eq:post_sim_scene}
  \vec{\mathcal{S}}^{\mathrm{sim}}
  =
  \{
  \vec{\mathcal{M}}_{i}^{\mathrm{init}},
  s_{i}^*,
  \mathbf{t}_{i}^{\mathrm{pre}}
  + \Delta \mathbf{t}^{\mathrm{pen}}_{i}
  + \Delta \mathbf{t}^{\mathrm{grv}}_{i},
  \Delta \mathbf{R}^{\mathrm{grv}}_{i} \mathbf{R}_{i}^{\mathrm{pre}}
  \}_{i=1}^N.
\end{equation}
Here, $\Delta \mathbf{R}^{\mathrm{grv}}_{i}$ is the identity matrix for objects tagged as \textit{free}. For wall-anchored objects, the constrained simulation returns the full pose change
$(\Delta \mathbf{t}_i^{\mathrm{grv}}, \Delta \mathbf{R}_i^{\mathrm{grv}})$ induced by rotation about the anchor.

\noindent\textbf{Displacement-Based Shape Correction Criterion.}
We use the physics-induced translation as a diagnostic signal for shape correction.
First, we compute the object's OBB (Oriented Bounding Box) in the pre-simulation scene $\vec{\mathcal{S}}^{\mathrm{pre}}$:
\begin{equation}\label{eq:obb_pre}
  \mathcal{B}^\mathrm{pre}_{i}
  =
  \operatorname{OBB}
  (
  \mathbf{R}_{i}^{\mathrm{pre}}
  \left(
  s_{i}^* \vec{\mathcal{M}}_{i}^{\mathrm{init}}
  \right)
  +
  \mathbf{t}_{i}^{\mathrm{pre}}).
\end{equation}
Let $\ell_{i}^{\mathrm{pre}}$ be the side length of $\mathcal{B}^\mathrm{pre}_{i}$ along the OBB axis most aligned with the unit gravity direction $\hat{\mathbf{g}}$. We then define the normalized gravity-axis displacement as
\begin{equation}\label{eq:stretch_ratio}
  \rho_i
  =
  \frac{
  \left| \hat{\mathbf{g}}^{\top}
  \left(
  \Delta \mathbf{t}^{\mathrm{pen}}_i +
  \Delta \mathbf{t}^{\mathrm{grv}}_i
  \right)
  \right|
  }{
  \ell_i^{\mathrm{pre}}
  }.
\end{equation}
$\rho_i$ measures how much each object in the post-simulation scene $\vec{\mathcal{S}}^{\mathrm{sim}}$ deviates from its counterpart in the pre-simulation scene $\vec{\mathcal{S}}^{\mathrm{pre}}$, which is the scene most closely aligned with the input image.
When $\rho_i$ is small, we regard the deviation as the result of minor errors accumulated through the pipeline and correct the scene with a lightweight adjustment.
In contrast, when $\rho_i$ is large, we attribute the deviation to a more fundamental shape-sampling failure, since our pipeline already performs substantial pose refinement. 
Specifically, if $\rho_i \geq 0.15$, we attribute the error to an unreliable SAM3D shape sample and trigger shape resampling. Otherwise, we correct the object by stretching it along the gravity-aligned OBB axis.

\subsection{Physics-Informed Shape Correction}\label{sec:shape_correction}

\noindent\textbf{Gravity-axis Stretch.}
When $\rho_i < 0.15$, we regard the discrepancy between $\vec{\mathcal{S}}^{\mathrm{pre}}$ and $\vec{\mathcal{S}}^{\mathrm{sim}}$ as a small accumulated error and correct it with a lightweight stretch rather than resampling, since resampling brings no measurable shape benefit in this regime (\tabref{tab:tau_main}).
Let $\hat{\mathbf{u}}_{i}^{\mathrm{grv}}$ be the OBB axis of $\mathcal{B}^{\mathrm{pre}}_i$ most aligned with the gravity direction $\hat{\mathbf{g}}$, and let
$\hat{\mathbf{a}}_{i}^{\mathrm{grv}} = (\mathbf{R}^{\mathrm{pre}}_i)^\top \hat{\mathbf{u}}_{i}^{\mathrm{grv}}$
be the corresponding axis in the canonical mesh frame.
We use the signed gravity-axis displacement
\begin{equation}\label{eq:stretch_displacement}
  \eta_i
  =
  \frac{
  \hat{\mathbf{g}}^\top
  \left(
  \Delta \mathbf{t}^{\mathrm{pen}}_i
  +
  \Delta \mathbf{t}^{\mathrm{grv}}_i
  \right)
  }{
  \ell_i^{\mathrm{pre}}
  }
\end{equation}
to stretch the canonical mesh along $\hat{\mathbf{a}}_{i}^{\mathrm{grv}}$.
The stretched mesh is then renormalized to preserve the canonical-mesh convention, with the normalization absorbed into the object scale. The translation is updated so that the OBB face opposite to gravity remains fixed.
This yields the stretched object
$(\vec{\mathcal{M}}^{\mathrm{str}}_i, s_i^{\mathrm{str}}, \mathbf{t}_i^{\mathrm{str}}, \Delta \mathbf{R}^{\mathrm{grv}}_{i}\mathbf{R}_i^{\mathrm{pre}})$.
See \secref{sec:supp_shape} for details.

\noindent\textbf{Amodal Shape Resampling.}
When $\rho_i \geq 0.15$, we attribute the error to an unreliable SAM3D shape sample, which typically occurs when severe occlusion causes SAM3D to fail at amodal shape completion.
We compute a desired OBB $\mathcal{B}^{\mathrm{str}}_{i}$ by stretching $\mathcal{B}^{\mathrm{pre}}_{i}$ along the gravity axis by the simulated displacement, and use it to guide shape resampling.
Since the true amodal mask is unavailable, we construct an auxiliary crop mask $\vec{M}'_i$ by augmenting the modal segmentation mask $\vec{M}_i$ with the projected lowest part of $\mathcal{B}^{\mathrm{str}}_{i}$, as shown in \figref{fig:pipeline}. 
Although $\vec{M}'_i$ is not a true amodal segmentation mask, it yields an amodal crop image $\vec{I}_{\vec{M}'_i}$ that covers the expected hidden extent of the target object.
We then run SAM3D with this physics-informed crop:
\begin{equation}\label{eq:resample}
  \Phi(\vec{I}, \vec{M}_i, \vec{I}_{\vec{M}'_i}, \vec{D})
  =
  (\vec{\mathcal{M}}_i^\mathrm{re}, s_i^\mathrm{re}, \tilde{\mathbf{t}}_i^\mathrm{re}, \mathbf{R}_i^\mathrm{re}).
\end{equation}
Since the resampled shape has different canonical geometry and scale, the previous scale and rotation are no longer valid, while the desired OBB $\mathcal{B}^{\mathrm{str}}_{i}$ provides the target object location.
We therefore keep the resampled mesh, scale, and rotation from SAM3D, but set the translation to the OBB center, $\mathbf{t}_i^\mathrm{re} := \operatorname{center}(\mathcal{B}^{\mathrm{str}}_{i})$.
The resulting resampled object is represented as
$(\vec{\mathcal{M}}_i^{\mathrm{re}}, s_i^{\mathrm{re}}, \mathbf{t}_i^{\mathrm{re}}, \mathbf{R}_i^{\mathrm{re}})$.

\noindent\textbf{SAM3D Fine-tuning for Amodal Resampling.}
To make SAM3D robust to the resampling input in \eqnref{eq:resample}, we fine-tune it with synthetic preference pairs that contrast occlusion-failed shape latents against edited full-object completion latents.

We use a DPO-style objective~\cite{dso,dpo,diffusion-dpo} that compares the flow-matching~\cite{flow-matching} losses of preferred and rejected shape latents:
\begin{equation}\label{eq:fm_dpo}
  \mathcal{L}_{\mathrm{FM\mbox{-}DPO}}(\theta)
  =
  \mathbb{E}_{i}
  \left[
  w_i
  \left(
  -\log \sigma
  \left(
  -\frac{\beta}{2}
  \left(
  d_{\theta,i} - d_{\mathrm{ref},i}
  \right)
  \right)
  \right)
  \right],
\end{equation}
where $w_i$ is the sample weight, $\sigma$ is the sigmoid function, $\beta$ controls the preference strength, $d_{\theta,i}$ is the win--lose flow-matching loss difference of the current model, and $d_{\mathrm{ref},i}$ is the corresponding difference computed by the frozen base SAM3D reference model.
This objective encourages the fine-tuned model to assign a lower relative flow-matching loss to the completed amodal latent than to the occlusion-failed latent.
To also directly supervise the completed amodal latent, we add a supervised fine-tuning (SFT) term:
\begin{equation}\label{eq:total_loss}
  \mathcal{L}_{\mathrm{total}}(\theta)
  =
  \mathcal{L}_{\mathrm{FM\mbox{-}DPO}}(\theta)
  +
  \lambda_{\mathrm{SFT}}\mathcal{L}_{\mathrm{SFT}}(\theta).
\end{equation}
Instead of updating the full SAM3D generator, we train only LoRA~\cite{lora} adapters inserted into the attention projections of the shape generation branch.
The LoRA adapters are trained once offline and kept fixed for all test scenes, so amodal resampling runs feed-forward without gradient-based optimization or per-scene fine-tuning.
Details of preference-pair construction, loss definitions, and training weights are provided in \secref{sec:supp_shape}.

\subsection{Final Composition}\label{sec:final}

The preceding stages produce corrected object states, but they do not yet define the final simulation-ready layout: the diagnostic simulation in \secref{sec:physics} is designed to extract clean correction signals, whereas the final scene must be physically settled under the object's full pose constraints.
We first assemble each object from its corrected state. Wall-anchored objects retain the initial geometry and use the constrained pose obtained from diagnostic simulation:
$(\vec{\mathcal{M}}_{i}^{\mathrm{init}}, s_{i}^*,
\mathbf{t}_{i}^{\mathrm{pre}}+\Delta \mathbf{t}^{\mathrm{pen}}_{i}+\Delta \mathbf{t}^{\mathrm{grv}}_{i},
\Delta \mathbf{R}^{\mathrm{grv}}_{i}\mathbf{R}_{i}^{\mathrm{pre}})$.
Free objects are replaced by either the stretched state
$(\vec{\mathcal{M}}^{\mathrm{str}}_i, s_i^{\mathrm{str}},
\mathbf{t}_i^{\mathrm{str}}, \Delta \mathbf{R}^{\mathrm{grv}}_{i}\mathbf{R}_i^{\mathrm{pre}})$
or the resampled state
$(\vec{\mathcal{M}}_i^{\mathrm{re}}, s_i^{\mathrm{re}},
\mathbf{t}_i^{\mathrm{re}}, \mathbf{R}_i^{\mathrm{re}})$,
according to the correction branch selected in \secref{sec:physics}.
Starting from these assembled states, we rerun sequential penetration resolution to remove residual overlaps introduced by shape correction.
Unlike the diagnostic probe, the final simulation does not freeze free objects at first contact.
Instead, free objects are simulated as full 6-DoF rigid bodies, while \textit{point-anchored} and \textit{line-anchored} objects follow their corresponding rotational constraints.
We let the scene settle under gravity for a fixed duration and take the resulting configuration as the final simulation-ready compositional 3D scene $\vec{\mathcal{S}}$.

%% file: figures/30_pipeline/figure.tex
\begin{figure*}[t!]
  \centering
  \includegraphics[width=\textwidth]{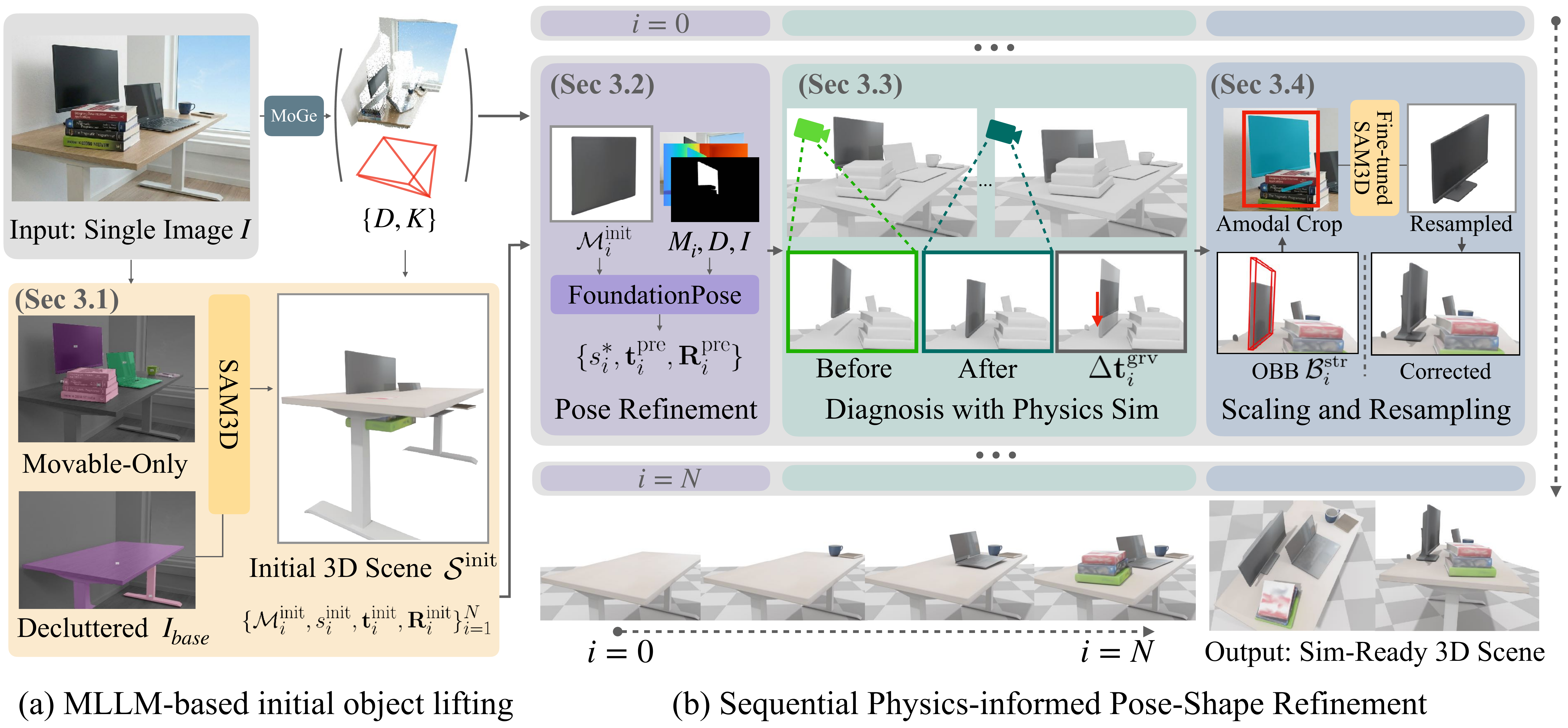}
  \caption{\textbf{Pipeline Overview.}
(a)~From a single image $I$, foundation priors decompose the scene into object meshes with poses.
(b)~Each object then passes through pose initialization, diagnostic physics simulation, and shape correction. Completed objects are frozen as colliders, and physics displacements drive the correction. The corrected objects settle into a simulation-ready 3D scene.
}
\label{fig:pipeline}
  \vspace{-10pt}
\end{figure*}

%% file: sections/04_result.tex
\section{Experiments}\label{sec:experiments}

\subsection{Experimental Setup}\label{sec:exp_setup}

\input{figures/40_result1/figure}

\input{tables/040_quant1}

\noindent\textbf{Implementation Details.}
For physical simulation and evaluation, we employ MuJoCo~\cite{2012mujoco} and extract collision geometries for each reconstructed object using V-HACD.
For SAM3D fine-tuning, we use 13,926 latent preference pairs, with 6,866 occlusion-completion pairs and 7,060 clean preservation pairs.
The pair weights $w_i$ in \eqnref{eq:fm_dpo} are set to $1.0$ and $0.1$, respectively, and we use $\beta=500$ and $\lambda_{\mathrm{SFT}}=300$.
We freeze the pretrained SAM3D model as $\theta_{\mathrm{ref}}$ and fine-tune only shape-branch LoRA~\cite{lora} adapters in the Stage-1 MM-DiT backbone, applying the flow-matching loss only to the Stage-1 shape latent.
The adapters use rank $r=64$, scaling $\alpha=128$, no dropout, and are applied to 120 linear layers across 24 transformer blocks.
We train with AdamW~\cite{loshchilov2017adamw} using a global batch size of 16 on a single NVIDIA RTX PRO 6000 Blackwell GPU with four gradient accumulation steps, a learning rate of $1\times10^{-5}$ after 300 warmup steps, weight decay $0.01$, gradient clipping at $1.0$, and bfloat16 mixed precision for 5,000 steps.

\input{figures/43_ablation1/figure}

\noindent\textbf{Datasets.}
We evaluate multi-object composition and physical stability on three complementary datasets.
\begin{itemize}[leftmargin=*, nosep]
    \item \textbf{GraspClutter6D}~\cite{back2025graspclutter6d} provides 94 real cluttered scenes in its YCB-V test split. For each scene, we pick the view with the largest margin between visible objects and the image boundary to avoid truncation, use the provided visible masks as ground truth, and segment unannotated supports such as tables, boxes, and shelves with SAM3.
    \item \textbf{Aria Digital Twin (ADT)}~\cite{pan2023ariadigitaltwin} supplies per-instance 3D ground-truth meshes and 6-DoF poses. We sample $40$ static scenes without humans or moving objects (\secref{sec:11}) to evaluate reconstruction, pose, and full-scene stability.
    \item \textbf{\dataname} contains $50$ synthetic images of challenging, out-of-distribution everyday layouts generated from text and image prompts~\cite{gemini}.
\end{itemize}

\noindent\textbf{Evaluation Metrics.}
Visual fidelity alone is insufficient if scenes collapse under physics, while trivial stability (e.g., naively projecting objects onto the floor) destroys the original spatial intent. We therefore compute all metrics \emph{post-simulation}, after objects settle under gravity for $5$\,s in MuJoCo~\cite{2012mujoco}. Measuring alignment after settling penalizes both physical collapse and distorted layouts, making it more relevant to downstream simulation than alignment before settling.
\begin{itemize}[leftmargin=*, nosep]
  \item \textbf{Residual motion} ($D_{\text{mean}}$, $E_{\text{pk}}$). Mean Displacement is the average translation of each object between its reconstructed and settled poses, and Peak Energy is the peak kinetic energy of each object during simulation, averaged over objects. A simulation-ready scene should already be at rest, so residual motion indicates an unstable layout. Peak Energy further captures violent transients, such as overlapping bodies pushed apart at high speed, which net displacement does not reflect.
  \item \textbf{Penetration} (Penetr.). The Penetration Ratio is the fraction of objects with any post-simulation contact deeper than $2$\,mm. A penetrating object is not directly interactable, since collisions between overlapping bodies produce unnatural behavior.
  \item \textbf{View alignment} (ABO, ABO$_{\mathrm{fh/fo}}$). Average Best Overlap measures fidelity to the input view after settling by averaging each ground-truth object's highest IoU with any prediction, using 3D bounding boxes for ADT and 2D masks otherwise. A high standard ABO alone does not establish simulation usability, because it also credits penetrating objects that are stuck and unusable in simulation. We thus introduce ABO$_{\mathrm{fo}}$ and ABO$_{\mathrm{fh}}$, which assign zero to penetrating objects. ABO$_{\mathrm{fo}}$ scores only free objects, which are fully interactable, and ABO$_{\mathrm{fh}}$ adds hanging (wall-mounted) objects, which are partially interactable since they can only rotate about their anchors. We use their average ABO$_{\mathrm{fh/fo}}$, which weights fully interactable objects more heavily since free objects appear in both terms, while penalizing stuck, non-interactable objects.
\end{itemize}
Detailed formulations are given in \secref{sec:12}, and appearance metrics (PSNR, LPIPS, and CLIP distance) are reported in \secref{sec:supp_appearance}, where our method also performs best.

\input{figures/45_dso_tau/figure}

\noindent\textbf{Baselines.}
We compare against methods capable of input-view aligned compositional reconstruction, namely SAM3D~\cite{sam3d}, Gen3DSR~\cite{ardelean2025gen3dsr}, 3D-RE-GEN~\cite{sautter20253dregen}, and 3D-Fixer~\cite{3D-Fixer}. All methods are evaluated with identical MuJoCo parameters and the same kinematic DoF tags to prevent unfair simulation penalties for hanging objects. Because flawed baseline backgrounds (e.g., missing floors) often trigger immediate simulation collapse, we evaluate Gen3DSR and 3D-RE-GEN twice, once with their native backgrounds and once with our wall/floor base mesh replacing those backgrounds (\emph{our bg}), to isolate object reconstruction quality.

\subsection{Experimental Results}
\noindent\textbf{Quantitative Comparison.}
As shown in \tabref{tab:main_quant}, our method achieves state-of-the-art results across all datasets, yielding the highest ABO$_{\mathrm{fh/fo}}$ on every dataset. While Gen3DSR~\cite{ardelean2025gen3dsr} records a deceptively high standard ABO, its severe penetration reveals physically fused objects, causing its ABO$_{\mathrm{fh/fo}}$ to drop sharply. Furthermore, since baselines often lack physically accurate environments, we evaluate them by supplying our extracted boundary constraints (\emph{our bg}). Interacting with these proper solid boundaries fully exposes their inherent instability. For 3D-RE-GEN, we replace its background mesh with our wall boundary while retaining its native floor parameters. Despite this setup, its objects still fail to settle stably and maintain high energy errors.

\noindent\textbf{Qualitative Results.}
Visual comparisons in \figref{fig:qual1} confirm these quantitative trends. As baselines frequently fuse objects together or embed them into the background, their high penetration ratios cause severe collisions during simulation, scattering objects out of bounds and resulting in completely missing meshes in their visualizations. In contrast, our pipeline places reconstructed objects in precise, static alignment with the input image, faithfully preserving the original spatial layout. Even heavily occluded items are accurately anchored to their exact locations with complete 3D geometry.

\subsection{Ablation Study}
We ablate our pipeline on \dataname{} by progressively adding components (\tabref{tab:ablation}, \figref{fig:ablation}). Pose alignment (\textbf{+Align.}) yields the largest gain by improving visual alignment and resolving deep interpenetrations. Rigid physics (\textbf{+Pen.+Grv.}) then stabilizes dynamics, but incorrect geometries persist. Physics-informed shape correction (\textbf{+Re.}) resolves these distortions, and amodal resampling replaces flawed meshes, yielding the lowest displacement error and highest alignment scores.

Beyond layout stability, we test whether our fine-tuned amodal resampler completes occluded geometry. For every object that triggers resampling, a VLM-based pairwise comparison~\cite{wu2023gpteval3d} judges which rendered mesh is the more plausible completion given the input. Our resampler achieves a substantially higher Elo score and win rate than raw SAM3D and stretching (\tabref{tab:mesh_qual}, \figref{fig:dso_ft}).

\noindent\textbf{Correction Threshold.}
We vary the resampling threshold $\tau$ (\secref{sec:physics}) on ADT and measure shape error with ADD-S$_{\mathrm{norm}}$, the ADD-S error after rigid ICP alignment to the GT mesh, normalized by the GT object diameter (\tabref{tab:tau_main}, \secref{sec:supp_tau}). For $\rho_i$ below our default $\tau=0.15$, stretching and resampling yield nearly identical shape error, while resampling substantially reduces the error for larger $\rho_i$. Resampling every object ($\tau=0$) thus brings no measurable shape gain over our two-tier rule but requires 47\% more resampling calls and time, so gravity-axis stretching serves as an efficiency mechanism. The ablation on GraspClutter6D (\secref{sec:supp_abl_real}) shows similar trends on real data.

\subsection{Applications}\label{sec:applications}
\input{figures/44_application/figure}

Our physically stable 3D reconstruction unlocks a single capability: the scene can be dropped into any simulator as a collection of separately addressable rigid bodies with consistent contact geometry, with no manual scene-authoring step.
\figref{fig:application} presents two downstream systems built on this invariant.

\noindent\textbf{Physics-Based Character Control.}
We load reconstructed scenes into a physics simulator and train a humanoid control policy following DeVI~\cite{devi}. As our scene provides physically stable assets, we are able to generate physics-based character motion including dexterous human-object interaction (HOI) in a novel scene by leveraging a video diffusion model as an HOI-aware motion planner.

\noindent\textbf{Robot-arm Manipulation}
We use our reconstructions as input to a closed-loop VLM agent~\cite{vlmpose} for text-guided 6D pose prediction, with execution carried out by GraspNet-based grasping~\cite{fang2020graspnet} and OMPL motion planning~\cite{sucan2012open}.
Whereas naive SAM3D leaves occluded geometry incomplete, with objects breaking away from their image-observed positions, our image-aligned reconstructions are object-complete and physically grounded, providing stable grasp targets and collision-free trajectories.
The reconstructed scenes therefore serve as a controllable test bed for text-guided robot-arm manipulation.

%% file: figures/40_result1/figure.tex
\begin{figure*}[t!]
  \centering
  \includegraphics[width=\textwidth]{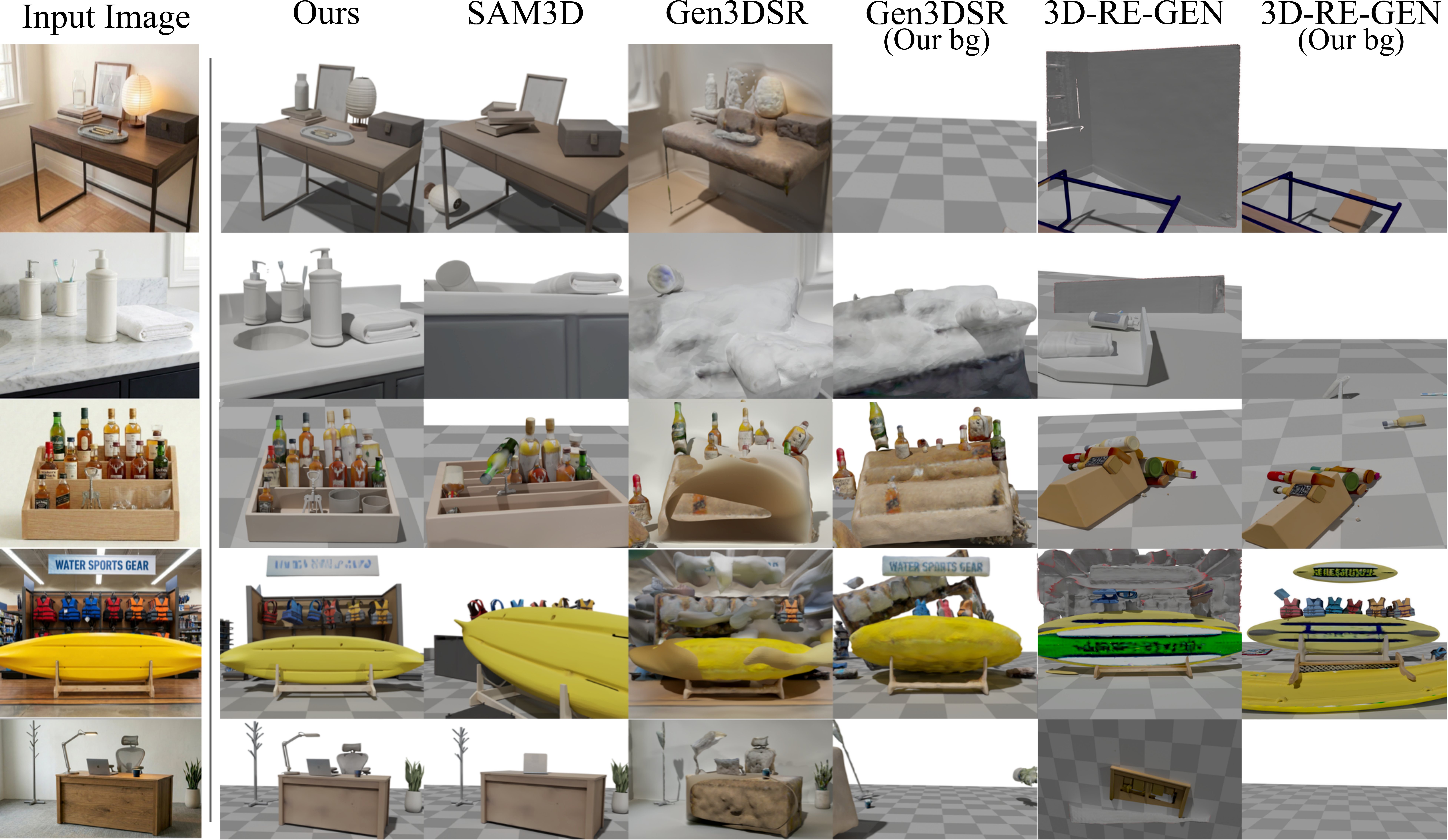}
    \caption{\textbf{Qualitative Comparison of \dataname.} 
    All scenes are visualized after gravity-driven physics simulation. Baselines either remain artificially stuck mid-air due to interpenetration or collapse catastrophically, while our method preserves both physical plausibility and input-view alignment in cluttered scenes.
    }
  \label{fig:qual1}
  \vspace{-8pt}
\end{figure*}

%% file: tables/040_quant1.tex
\definecolor{tabfirst}{rgb}{1,0.70,0.70}
\definecolor{tabsecond}{rgb}{1,0.85,0.70}
\definecolor{tabthird}{rgb}{1,1,0.70}

\begin{table*}[t]
  \centering
  \renewcommand{\arraystretch}{1.2}
  \setlength{\tabcolsep}{4pt}
  \resizebox{\textwidth}{!}{%
  \begin{tabular}{lccccccccccccccccc}
    \toprule
     & \multicolumn{5}{c}{GraspClutter6D} & & \multicolumn{5}{c}{AriaDigitalTwin} & & \multicolumn{5}{c}{GenWild} \\
    \cmidrule(lr){2-6} \cmidrule(lr){8-12} \cmidrule(lr){14-18}
    Method
      & ABO$_{\mathrm{fh/fo}}$ $\uparrow$
      & ABO $\uparrow$
      & $D_{\mathrm{mean}}$ $\downarrow$
      & $E_{\mathrm{pk}}$ $\downarrow$
      & Penetr.\ $\downarrow$
      & 
      & 3D ABO$_{\mathrm{fh/fo}}$ $\uparrow$
      & 3D ABO $\uparrow$
      & $D_{\mathrm{mean}}$ $\downarrow$
      & $E_{\mathrm{pk}}$ $\downarrow$
      & Penetr.\ $\downarrow$
      & 
      & ABO$_{\mathrm{fh/fo}}$ $\uparrow$
      & ABO $\uparrow$
      & $D_{\mathrm{mean}}$ $\downarrow$
      & $E_{\mathrm{pk}}$ $\downarrow$
      & Penetr.\ $\downarrow$ \\
    \midrule
    SAM3D~\cite{sam3d} & \cellcolor{tabthird}0.087 & 0.121 & 0.638 & 4.064 & \cellcolor{tabthird}24.5\% &  & 0.037 & 0.052 & 0.721 & \cellcolor{tabthird}3.556 & \cellcolor{tabsecond}16.4\% &  & \cellcolor{tabsecond}0.218 & \cellcolor{tabthird}0.263 & \cellcolor{tabthird}0.555 & 2.957 & \cellcolor{tabsecond}16.2\% \\
    Gen3DSR~\cite{ardelean2025gen3dsr} (our bg) & 0.014 & 0.036 & 0.880 & 5.644 & 50.3\% &  & \cellcolor{tabthird}0.070 & 0.095 & 0.932 & 5.769 & 17.2\% &  & 0.104 & 0.144 & 0.817 & 5.230 & \cellcolor{tabthird}16.9\% \\
    Gen3DSR~\cite{ardelean2025gen3dsr} & 0.014 & \cellcolor{tabsecond}0.219 & \cellcolor{tabsecond}0.267 & \cellcolor{tabsecond}1.587 & 75.8\% &  & 0.002 & \cellcolor{tabsecond}0.226 & \cellcolor{tabfirst}\textbf{0.091} & \cellcolor{tabfirst}\textbf{0.556} & 93.3\% &  & 0.046 & \cellcolor{tabsecond}0.477 & \cellcolor{tabsecond}0.255 & \cellcolor{tabsecond}1.214 & 72.7\% \\
    3D-RE-GEN~\cite{sautter20253dregen} (our bg) & 0.042 & 0.071 & 0.659 & 4.086 & 24.9\% &  & 0.044 & 0.060 & 0.835 & 4.261 & \cellcolor{tabthird}17.2\% &  & 0.110 & 0.154 & 0.612 & 3.491 & 19.7\% \\
    3D-RE-GEN~\cite{sautter20253dregen} & 0.056 & 0.100 & \cellcolor{tabthird}0.524 & \cellcolor{tabthird}2.790 & 31.4\% &  & 0.025 & 0.053 & 0.797 & 4.162 & 32.8\% &  & 0.104 & 0.164 & 0.608 & 2.790 & 27.0\% \\
    3D-Fixer~\cite{3D-Fixer} & \cellcolor{tabsecond}0.106 & \cellcolor{tabthird}0.122 & 0.606 & 3.455 & \cellcolor{tabfirst}\textbf{10.0\%} &  & \cellcolor{tabsecond}0.116 & \cellcolor{tabthird}0.159 & \cellcolor{tabthird}0.720 & 3.884 & 20.2\% &  & \cellcolor{tabthird}0.191 & 0.258 & 0.568 & \cellcolor{tabthird}2.680 & 20.1\% \\
    \midrule
    \textbf{Ours} & \cellcolor{tabfirst}\textbf{0.248} & \cellcolor{tabfirst}\textbf{0.267} & \cellcolor{tabfirst}\textbf{0.250} & \cellcolor{tabfirst}\textbf{1.092} & \cellcolor{tabsecond}10.3\% &  & \cellcolor{tabfirst}\textbf{0.247} & \cellcolor{tabfirst}\textbf{0.285} & \cellcolor{tabsecond}0.310 & \cellcolor{tabsecond}1.370 & \cellcolor{tabfirst}\textbf{8.8\%} &  & \cellcolor{tabfirst}\textbf{0.507} & \cellcolor{tabfirst}\textbf{0.532} & \cellcolor{tabfirst}\textbf{0.128} & \cellcolor{tabfirst}\textbf{0.420} & \cellcolor{tabfirst}\textbf{2.5\%} \\
    \bottomrule
  \end{tabular}%
  }
  \caption{\textbf{Main Quantitative Comparison.} ``(our bg)'' places each baseline's reconstructed objects in our reconstructed background (floor and walls). ABO$_{\mathrm{fh/fo}}$ averages the post-simulation ABO over non-penetrating free+hanging (wall-mounted) objects and over free-only objects, measuring the alignment of physically valid objects. Ours generally performs best and leads in ABO$_{\mathrm{fh/fo}}$ on all three datasets.}
  \label{tab:main_quant}
  \vspace{-14pt}
\end{table*}

%% file: figures/43_ablation1/figure.tex
\begin{figure*}[t!]
  \centering
  \begin{minipage}[h]{0.65\textwidth}
    \vspace{0pt}
    \centering
    \includegraphics[trim=0 0 0 0, clip, width=\linewidth]{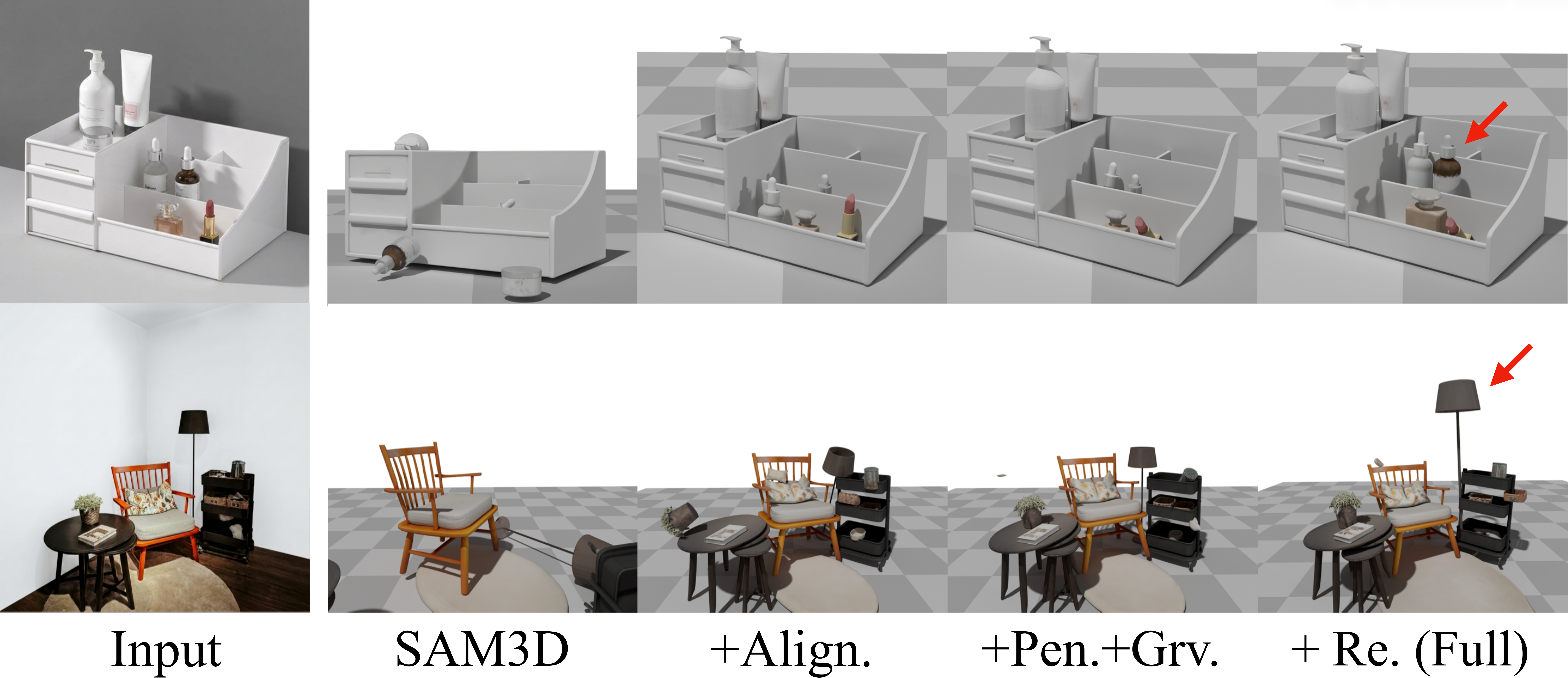}
    \captionof{figure}{\textbf{Qualitative Ablation.}
    Post-simulation rendering of 3D scene.
    From SAM3D, we progressively add per-object alignment (Align.),
    penetration resolution (Pen.), stabilized gravity simulation (Grv.),
    and shape resampling (Re.). The red arrows indicate that resampling successfully generates plausible, aligned shapes.}
    \label{fig:ablation}
  \end{minipage}%
  \hfill
  \begin{minipage}[h]{0.33\textwidth}
    \vspace{0pt}
    \centering
    \captionsetup{justification=RaggedRight,singlelinecheck=false}

\input{tables/042_quant_ablate}

    \input{tables/043_quant_meshqual}
  \end{minipage}
  \vspace{-10pt}
\end{figure*}

%% file: tables/042_quant_ablate.tex
\resizebox{\linewidth}{!}{%
\begin{tabular}{l|ccc}
  \toprule
    Variants
    & ABO $\uparrow$
    & ABO$_{\mathrm{fh/fo}}$ $\uparrow$
    & $D_{\mathrm{mean}}$ $\downarrow$ \\
  \midrule
  SAM3D & 0.263 & 0.218 & 0.555 \\
  \ +Align. & \cellcolor{tabthird}0.409 & 0.361 & \cellcolor{tabthird}0.364 \\
  \ +Pen. & 0.403 & \cellcolor{tabthird}0.372 & 0.395 \\
  \ +Grv. & \cellcolor{tabsecond}0.440 & \cellcolor{tabsecond}0.406 & \cellcolor{tabsecond}0.197 \\
  \midrule
  \ +Re. (Full) & \cellcolor{tabfirst}\textbf{0.532} & \cellcolor{tabfirst}\textbf{0.507} & \cellcolor{tabfirst}\textbf{0.128} \\
  \bottomrule
\end{tabular}%
}
\captionof{table}{\textbf{Quantitative Ablation.} Each stage follows \figref{fig:ablation}.}
\label{tab:ablation}

%% file: tables/043_quant_meshqual.tex
\centering

\resizebox{\linewidth}{!}{%
    \begin{tabular}{lccc}
        \toprule
        \textbf{Metric} & \textbf{Ours} & \textbf{SAM3D} & \textbf{Stretched} \\
        \midrule
        Elo $\uparrow$ & \cellcolor{tabfirst}\textbf{1083.4} & \cellcolor{tabsecond}931.5 & \cellcolor{tabthird}908.9 \\
        Win Rate (\%) $\uparrow$  & \cellcolor{tabfirst}\textbf{71.9} & \cellcolor{tabsecond}41.3 & \cellcolor{tabthird}36.8 \\
        \bottomrule
        \end{tabular}%
    }
\captionof{table}{\textbf{Mesh Completion Quality.} VLM-based evaluation~\cite{wu2023gpteval3d} on \dataname{}. Win Rate is the average pairwise win rate against the other candidates.}
\label{tab:mesh_qual}

%% file: figures/45_dso_tau/figure.tex
\begin{figure*}[t!]
  \centering
  \begin{minipage}[t]{0.60\textwidth}
    \vspace{0pt}
    \centering
    \captionsetup{justification=RaggedRight,singlelinecheck=false}
    \includegraphics[width=\linewidth]{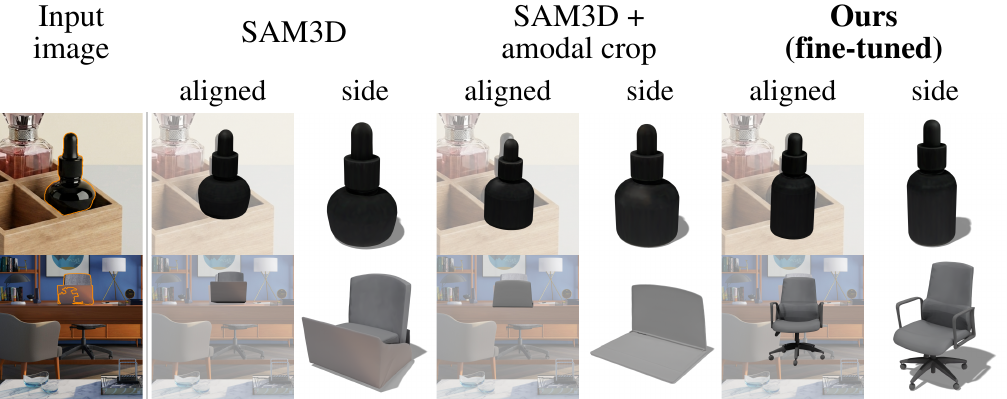}
    \captionof{figure}{\textbf{Effect of SAM3D Fine-tuning.} Under heavy occlusion, our resampling uses an amodal crop covering the hidden extent predicted by the gravity-stretched OBB. Only our fine-tuned model turns this crop into complete shapes aligned with the input view.}
    \label{fig:dso_ft}
  \end{minipage}%
  \hfill
  \begin{minipage}[t]{0.37\textwidth}
    \vspace{0pt}
    \centering
    \captionsetup{justification=RaggedRight,singlelinecheck=false}
    \footnotesize
    \setlength{\tabcolsep}{2pt}
    \begin{tabular}{lccc}
      \toprule
      $\tau$ & \#Resampled & ADD-S$_{\mathrm{norm}}$ $\downarrow$ & Time (s) \\
      \midrule
      $0$         & 708 & 0.0334 & 318.9 \\
      $0.15$      & 482 & \textbf{0.0333} & 217.1 \\
      $\infty$    & 0   & 0.0603 & 0.0 \\
      \bottomrule
    \end{tabular}
    \captionof{table}{\textbf{Resampling Threshold} on ADT. $\tau{=}0$ resamples every object and $\tau{=}\infty$ only stretches. Our default $\tau{=}0.15$ is the most economical setting, matching the shape error of $\tau{=}0$ while cutting resampling calls and time by $32\%$.}
    \label{tab:tau_main}
  \end{minipage}
  \vspace{-10pt}
\end{figure*}

%% file: figures/44_application/figure.tex
\begin{figure*}[t!]
  \centering
  \begin{minipage}[t]{0.48\textwidth}
    \centering
    \includegraphics[width=\linewidth]{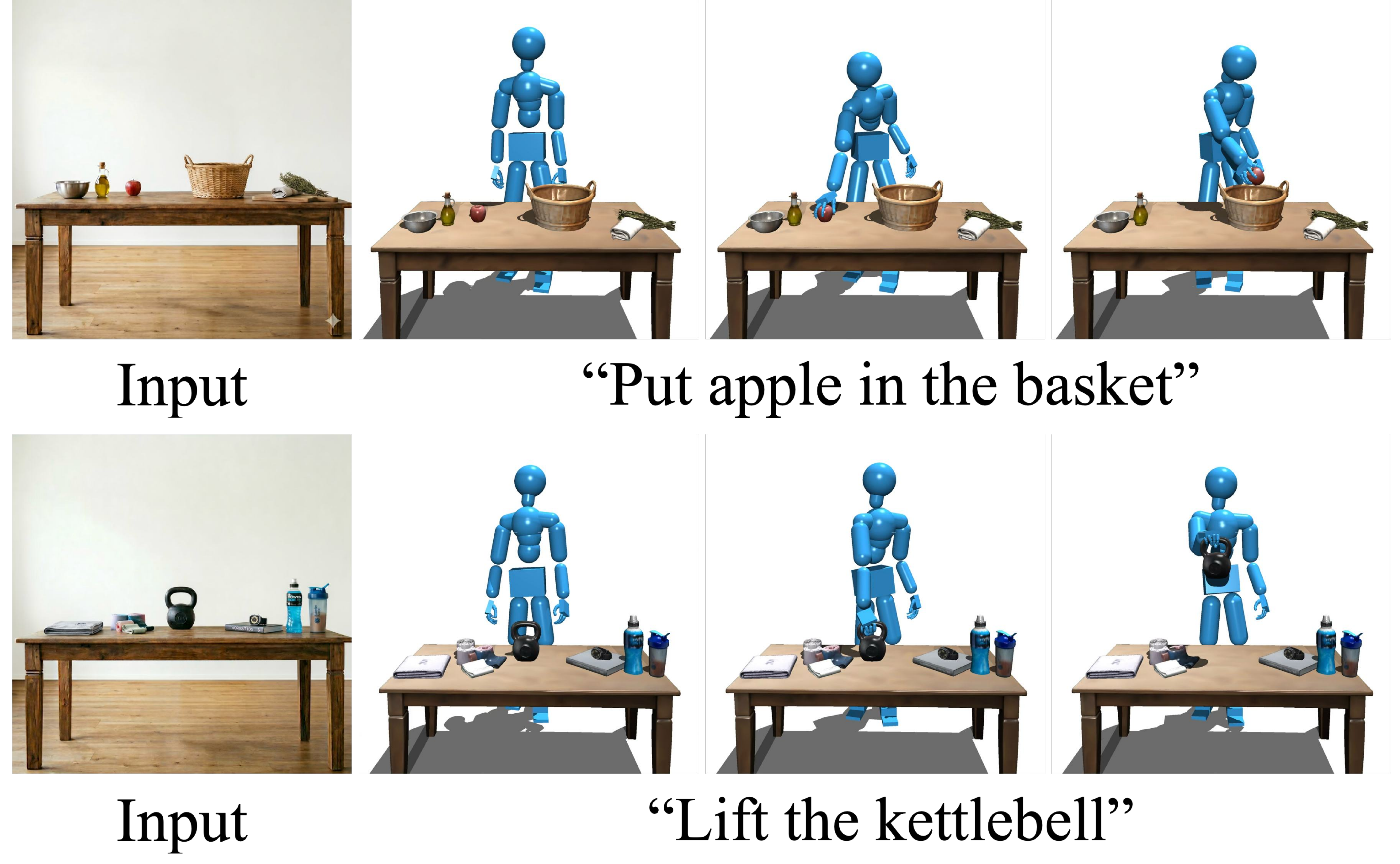}\\[2pt]
    \small (a) Physically plausible HOI learning
  \end{minipage}%
  \hfill
  \begin{minipage}[t]{0.50\textwidth}
    \centering
    \includegraphics[width=\linewidth]{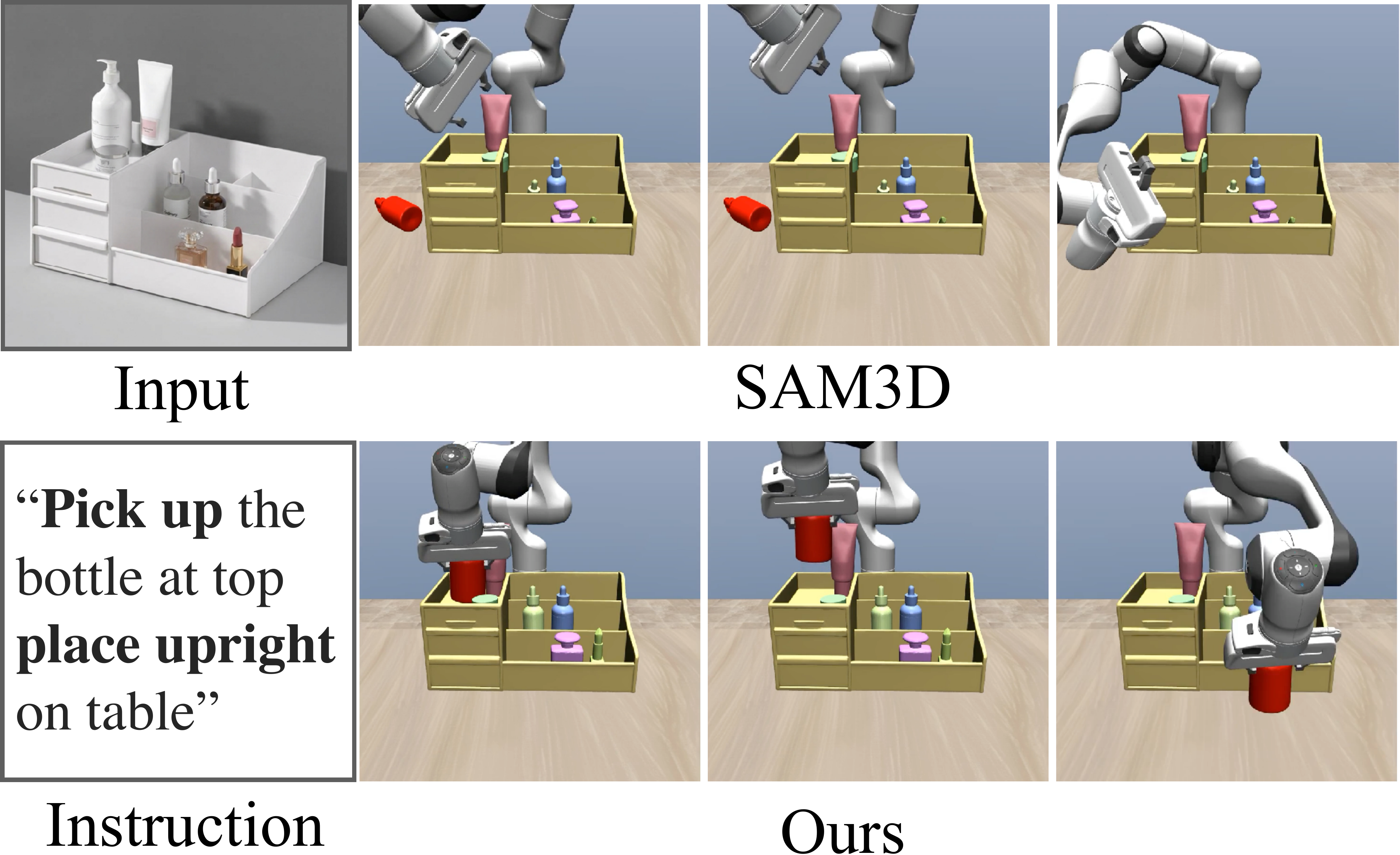}\\[2pt]
    \small (b) Robotic manipulation test
  \end{minipage}
  \caption{\textbf{Applications.} Two downstream uses of our object-complete, physics-stable reconstructions: (a) physics-based character control~\cite{devi}, (b) cluttered robotic manipulation~\cite{vlmpose}.}
  \label{fig:application}
  \vspace{-10pt}
\end{figure*}

%% file: sections/05_discussion.tex
\section{Discussion}\label{sec:discussion}
We reframe physical simulation from a post-hoc validator into an in-the-loop diagnostic signal that inverts physical violations into geometric corrections, constraining the structural ambiguities of monocular perception. The resulting scenes are directly usable in downstream tasks such as robotic manipulation and reinforcement learning. While our sequential protocol cannot revise early estimates from later evidence, integrating physical dynamics as a generative supervisory signal offers a scalable foundation for reconstructing simulation-ready scenes from single images, with joint scene-level optimization as a natural next step.

\noindent\textbf{Limitations.}
Our pipeline composes several pretrained modules (segmentation, monocular depth, single-image lifting, pose refinement, and VLM queries), so upstream errors can propagate. Pose initialization is a key bottleneck (\secref{sec:supp_upstream}).
We assume rigid objects under rigid-body contact, so residuals for articulated or deformable objects may not reflect their true degrees of freedom.
The sequential protocol is order-dependent and cannot revise earlier objects from later evidence, and absolute scale remains tied to monocular depth, since gravity-axis residuals reveal support errors rather than metric scale.
A more compact formulation that jointly optimizes shape and layout with fewer modules is a natural next step.

%% file: sections/10_supple.tex
\clearpage
\appendix

\input{figures/100_supp0/figure}

\input{tables/100_supp_quant1}

\section{Implementation Details for the Method}\label{sec:method_supp}

This appendix expands the technical details that were deferred from \secref{sec:method}.

\subsection{Preprocessing Pipeline}\label{sec:supp_pp}
\noindent\textbf{Decluttered image generation.}
The base structures in \secref{sec:base_only} are selected per category, so a base structure may consist of one or several object instances. The decluttered image is most beneficial in complex scenes and can be omitted for simple scenes with few objects and little occlusion.

\noindent\textbf{Quality gate for base-only image generation.}
A VLM checks each base-only image $I_\text{base}$ produced by the MLLM-based editor for removal completeness and edge alignment. Removal completeness checks whether all foreground objects have been removed. For edge alignment, we overlay edges extracted from $I_\text{base}$ onto the input image $I$ and ask the VLM whether they align with the base structure in $I$. This verifies that the base structure is preserved and stays in place. Generation and verification are repeated until both criteria are satisfied.

\noindent\textbf{Cross-path de-duplication via DINOv3.}
The two extraction paths in \secref{sec:base_only} can reconstruct the same support structure. SAM3 on $I$ can segment it (e.g., a desk edge) as a placed-object candidate, while SAM3D on $I_\text{base}$ already reconstructs it as part of the base structure mesh.
To detect this, we compare dense DINOv3~\cite{simeoni2025dinov3} feature maps $\mathbf{F}, \mathbf{F}_{\text{base}} \in \mathbb{R}^{H \times W \times C}$ of $I$ and $I_\text{base}$ through the per-pixel cosine difference
\begin{equation}\label{eq:dino_diff}
  \Delta(\mathbf{p}) = 1 - \frac{\mathbf{F}(\mathbf{p}) \cdot \mathbf{F}_{\text{base}}(\mathbf{p})}{\|\mathbf{F}(\mathbf{p})\|\;\|\mathbf{F}_{\text{base}}(\mathbf{p})\|}.
\end{equation}
Thresholding $\Delta$ at $\delta_\text{dino}$ yields a binary edited-region map $P = \mathbb{1}[\Delta > \delta_\text{dino}]$ that marks the pixels of $I$ removed in $I_\text{base}$.
We retain a placed-object candidate mask $M_j$ only if
\begin{equation}\label{eq:mask_overlap}
  |M_j \cap P|\,/\,|M_j| > \alpha_\text{mask},
\end{equation}
so that candidates outside the edited region, which the base path has already lifted, are dropped.
\figref{fig:instance_extraction} summarizes this pipeline.

\noindent\textbf{Floor and wall plane extraction with Manhattan rectification.}
We back-project the SAM3 floor and wall masks to 3D with metric depth and fit planes with RANSAC, using residual thresholds of $5$\,cm for the floor and $1$\,cm for each wall. We keep up to four walls with the largest coverage whose orientations are mutually parallel or orthogonal. The up direction is given by the cross product of two orthogonal walls, or by the floor normal when no such pair exists, and all wall normals are rectified to be perpendicular to it. After object fitting, we re-estimate the planes with the fitted meshes, which provide a stronger up-direction signal than depth alone, and snap the floor offset to the lowest vertex among all fitted meshes to close the gap between floor-standing objects and the floor. Tables and desks are further refined as described in \secref{sec:tabletop_refine}.

\subsection{Pose Alignment Details}

\noindent\textbf{Translation--Scale Initial Alignment.}
\label{sec:initial_alignment}
Before applying FoundationPose~\cite{wen2024foundationpose} in Eq.~\ref{eq:fdp_refine}, we perform a lightweight translation--scale alignment for each SAM3D object. 
Although FoundationPose is effective as a local pose refiner, we find that it is sensitive to large errors in the initial scale or object center: when the rendered object is substantially misaligned with the observed mask and depth, the render-and-compare refinement can converge to an incorrect
pose or fail to recover. 
We therefore use the masked depth observation to bring the SAM3D prediction into a reasonable basin before FoundationPose.
Given the canonical mesh $\vec{\mathcal{M}}_i^{\mathrm{init}}$, initial prediction $(\vec{R}_i^{\mathrm{init}}, \vec{t}_i^{\mathrm{init}}, s_i^{\mathrm{init}})$, and the target point cloud $\vec{\mathcal{P}}_i$ back-projected from the masked depth map, we solve only for the isotropic scale and translation $(s_i,\vec{t}_i)$ while keeping $\vec{R}_i^{\mathrm{init}}$ fixed. 
We avoid optimizing rotation in this stage because chamfer-style point-set losses provide weak and often degenerate rotation signals under monocular occlusion, partial visibility, and symmetric or textureless geometry. This step therefore serves as a 4-DoF initialization that corrects scale and centering errors before the rotation-aware refinement.

Since SAM3D meshes are complete while the input depth observes only the visible surface, directly matching all mesh vertices to $\vec{\mathcal{P}}_i$ biases the alignment toward self-occluded geometry. We therefore rasterize $\vec{\mathcal{M}}_i^{\mathrm{init}}$ once at the initial pose and keep only the visible vertex subset $\mathcal{V}_i^{\mathrm{vis}}$. The translation--scale alignment is then optimized with a one-sided chamfer objective from the observed point cloud to the visible mesh vertices:
\begin{equation}
\mathcal{L}_{\mathrm{align}}(s_i,\vec{t}_i) = \frac{1}{|\vec{\mathcal{P}}_i|} \sum_{\vec{y}\in \vec{\mathcal{P}}_i} \min_{\vec{v}\in\mathcal{V}_i^{\mathrm{vis}}} \left\| \vec{y} - \left(s_i \vec{R}_i^{\mathrm{init}}\vec{v} + \vec{t}_i \right) \right\|_2 .
\label{eq:align_loss}
\end{equation}

The asymmetric direction encourages the predicted visible surface to cover the observed depth points, while avoiding penalties from unobserved mesh regions caused by self-occlusion or inter-object occlusion.

However, the one-sided loss alone can favor over-grown meshes, since a large mesh may still cover all observed points. To reduce this failure mode, we run multi-start Adam optimization from several initial scales $s_{i,0}^{(k)}=\rho_k s_i^{\mathrm{init}}$, with $\rho_k\in\{0.5,0.75,1.0,1.25\}$, and apply a weak log-scale prior:
\begin{equation}
\mathcal{L}^{(k)} = \mathcal{L}_{\mathrm{align}}(s_i,\vec{t}_i) + \lambda \left( \log s_i - \log s_i^{\mathrm{init}} \right)^2 .
\label{eq:align_reg}
\end{equation}

After convergence, we rank the candidates using a combination of silhouette recall and a weakly weighted bidirectional chamfer score:
\begin{equation}
k^*=\arg\max_k\frac{1}{2}\widetilde{\mathrm{Recall}}^{(k)} + \frac{1}{2}\left(1 - \widetilde{\mathcal{L}}_{\mathrm{bi}}^{(k)}\right),
\label{eq:align_select}
\end{equation}

where the recall term penalizes under-coverage of the input mask, and the bidirectional chamfer term discourages excessive scale growth. The selected $(s_i^*,\vec{t}_i^*)$ is then used as the initialization for FoundationPose, while the rotation remains initialized by $\vec{R}_i^{\mathrm{init}}$.

\noindent\textbf{Tabletop-Plane Refinement.}
\label{sec:tabletop_refine}
For tables and desks, even small residual tilt or height errors on the top support surface produce visible floating or penetration artifacts once small objects are placed on top. We therefore refine them after FoundationPose by aligning the mesh top surface to a tabletop plane detected from MoGe depth. 

Given the post-refinement mesh and the scene up direction $\mathbf{u}$, we first extract a depth-supported tabletop plane $(\mathbf{n}^d,\mathbf{p}^d)$ by
filtering upward-facing pixels within the object mask, applying a height-band constraint to suppress clutter and under-shelf regions, and fitting a plane with RANSAC. We also estimate the mesh top plane $(\mathbf{n}^m,\mathbf{p}^m)$ from canonical top vertices identified along the transformed up direction. The mesh top is then rigidly snapped onto the detected support plane by aligning normals and removing the residual height gap:

\begin{equation}
\label{eq:hard_snap_short}
    T_{\text{snap}}(v)=\vec{R}^{\mathrm{snap}}\,v + (\mathbf{p}^m - \vec{R}^{\mathrm{snap}}\mathbf{p}^m)+\delta,
\end{equation}

where $\vec{R}^{\mathrm{snap}}$ rotates $\mathbf{n}^m$ onto $\mathbf{n}^d$, and $\delta = \bigl((\mathbf{p}^d-\mathbf{p}^m)\cdot\mathbf{n}^d\bigr)\,\mathbf{n}^d$ translates the mesh along the detected plane normal to close the residual gap to the support plane.

Starting from the snapped pose, we further optimize four residual degrees of freedom consisting of yaw rotation around the detected plane normal, two in-plane translations, and an isotropic scale. The optimization minimizes a symmetric chamfer between the rendered visible mesh point cloud and the depth-lifted observation, regularized toward the snapped initialization:
\begin{equation}
\label{eq:table_refine_short}
\min_{\theta,t_x,t_y,\log s}\;
\tfrac12\bigl(d_{\hat P \rightarrow P} + d_{P \rightarrow \hat P}\bigr) + \lambda\Bigl[(\theta/\theta_0)^2 + (t_x/t_0)^2 + (t_y/t_0)^2 + (\log s/\sigma_0)^2 \Bigr].
\end{equation}

Refinements with insufficient plane support, non-horizontal detected planes, or floor-plane degeneracies are discarded, and the FoundationPose result is kept.

\subsection{Gravity-Based Diagnostics Details}\label{sec:supp_sim}

\input{figures/102_wall_hang_supp/figure}
\noindent\textbf{How VLM tags object-specific pose DoF.}
Indoor scenes routinely contain objects that are not floor-supported, such as picture frames, wall-mounted TVs, and pendant lamps. Without special handling, the physics simulator would drop them to the ground and corrupt the reconstructed layout.
To handle these, we compose a two-panel query image from the input image for each object, highlighting the target on the left and isolating it against a desaturated background on the right, and ask a VLM to classify it as standing or hanging. Only objects with their own attachment to a wall or ceiling count as hanging, so an object resting on a wall-mounted shelf remains standing while the shelf is hanging.
For hanging objects we additionally elicit a pin count, which the simulator uses to anchor the object with the appropriate number of constraints. The count is one for freely swinging mounts (hooks, pendants) and two for rigid mounts (frames, brackets). 
The full prompt and qualitative examples are provided in \figref{fig:wall_hang_supp}.

\noindent\textbf{Wall-anchoring Mechanism.}
For each object $o$ flagged as \emph{hanging} by the VLM with pin count $K\!\in\!\{1,2\}$, we attach it to the reconstructed scene wall through $K$ ball-joint anchors. 
Each anchor pairs a point $\mathbf{p}_o^{(k)}$ on the object's surface with a point $\mathbf{p}_W^{(k)}$ on the wall surface, and the simulator keeps the two points coincident throughout the rollout.

\textbf{(1)~Choosing the object-side pin.} 
We restrict pin candidates to vertices that are simultaneously near the top of the object (along the anti-gravity direction) and on its wall-facing back side, excluding front- or bottom-facing vertices that would otherwise yield an inverted or wall-detached hang. 
For $K\!=\!1$ (hooks, pendant lamps, and other freely swinging mounts), we take the highest and most laterally centered vertex of this top-back region. For $K\!=\!2$ (picture frames, mounted TVs, and other wall-mounted objects), we take the leftmost and rightmost vertices of this top-back region along the wall-parallel horizontal direction. The object can rotate only about the axis joining the two anchors, so gravity keeps its face flat against the wall.

\textbf{(2)~Pairing each pin with a wall point.}
Each object-side pin is paired with its closest point on the wall mesh, which is the foot of the perpendicular dropped from the pin onto the wall surface. 
This preserves the natural standoff distance between the object and the wall rather than collapsing the object onto the wall plane.

\textbf{(3)~Force coupling between the two points.}
Each pin pair is realized as a MuJoCo ball-joint equality constraint that keeps the two points coincident at every simulation step. The constraint is enforced as a stiff, critically damped spring rather than a hard weld, so contact transients are absorbed without instability.

\noindent\textbf{Physics-mode parameters.}
The simulator runs at a timestep of $1/480$\,s with inelastic, stiff contacts, and the sliding and torsional friction coefficients are $1.0$ and $0.95$, respectively.

\begin{figure}[h]
    \centering
    \includegraphics[width=\textwidth]{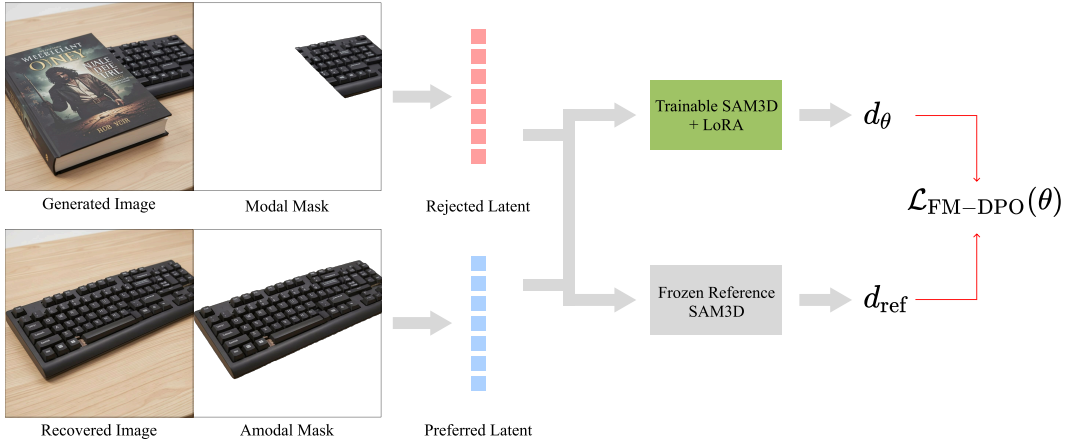}
    \caption{\textbf{SAM3D Fine-tuning Pipeline.} We fine-tune SAM3D for amodal shape resampling using synthetic preference pairs that contrast occlusion-failed latents with completed amodal latents. A flow-matching DPO objective with an additional preferred-sample SFT term updates only LoRA adapters while keeping the reference SAM3D frozen, and the resulting model is used at test time for physics-informed resampling.
    }
    \label{fig:training_pipeline}
\end{figure}

\noindent\textbf{Penetration Resolution.}
For each newly placed object, we resolve its overlap with the floor, the rectified walls, and every previously placed object while keeping it as the only dynamic body. Floor and wall contacts take priority over contacts with previously placed objects. At each iteration, the object is translated along the depth-weighted average normal of the highest-priority contacts by their maximum penetration depth plus a $7$\,mm margin. Objects that remain in contact after $200$ iterations or whose cumulative displacement exceeds $0.3$\,m skip the gravity-based diagnostics and shape correction, and are retained in the final scene.

\subsection{Shape Correction Details}\label{sec:supp_shape}

\noindent\textbf{Gravity-axis Stretch Details.}
For objects corrected by gravity-axis stretching, we use the signed displacement $\eta_i$ from \eqnref{eq:stretch_displacement}.
We orient the gravity-aligned OBB axis such that
$(\hat{\mathbf{u}}_{i}^{\mathrm{grv}})^\top \hat{\mathbf{g}} \geq 0$, and define its canonical-frame counterpart as
\begin{equation}
  \hat{\mathbf{a}}_{i}^{\mathrm{grv}}
  =
  (\mathbf{R}^{\mathrm{pre}}_i)^\top
  \hat{\mathbf{u}}_{i}^{\mathrm{grv}} .
\end{equation}
The stretch factor is
\begin{equation}
  \lambda_i = 1 + \eta_i .
\end{equation}
We first apply anisotropic stretching in the canonical mesh frame:
\begin{equation}\label{eq:supp_stretch_function}
  \bar{\vec{\mathcal{M}}}^{\mathrm{str}}_i
  =
  \mathrm{Stretch}_{\hat{\mathbf{a}}_{i}^{\mathrm{grv}}}
  \left(
  \vec{\mathcal{M}}^{\mathrm{init}}_i;\,
  \lambda_i
  \right),
\end{equation}
where $\mathrm{Stretch}_{\hat{\mathbf{a}}}(\vec{\mathcal{M}};\lambda)$ anisotropically scales the mesh by factor $\lambda$ along axis $\hat{\mathbf{a}}$ while leaving the orthogonal directions unchanged.
For a vertex $\mathbf{v}$ of a canonical mesh centered at the origin, this operation is
\begin{equation}
  \mathbf{v}'
  =
  \lambda_i
  \left(
  \mathbf{v}^{\top}\hat{\mathbf{a}}_{i}^{\mathrm{grv}}
  \right)
  \hat{\mathbf{a}}_{i}^{\mathrm{grv}}
  +
  \left(
  \mathbf{I}
  -
  \hat{\mathbf{a}}_{i}^{\mathrm{grv}}
  (\hat{\mathbf{a}}_{i}^{\mathrm{grv}})^\top
  \right)
  \mathbf{v}.
\end{equation}

Let $\gamma_i$ be the longest side length of the AABB of
$\bar{\vec{\mathcal{M}}}^{\mathrm{str}}_i$.
To preserve the canonical-mesh convention, we renormalize the stretched mesh and absorb the normalization factor into the object scale:
\begin{equation}\label{eq:supp_stretch_mesh_cano}
  \vec{\mathcal{M}}^{\mathrm{str}}_i
  =
  \frac{1}{\gamma_i}
  \bar{\vec{\mathcal{M}}}^{\mathrm{str}}_i,
  \qquad
  s_i^{\mathrm{str}}
  =
  \gamma_i s_i^* .
\end{equation}
Finally, we update the translation so that the OBB face opposite to gravity remains fixed:
\begin{equation}\label{eq:supp_stretch_transl}
  \mathbf{t}_i^{\mathrm{str}}
  =
  \mathbf{t}_i^{\mathrm{pre}}
  +
  \frac{1}{2}
  \eta_i
  \ell_i^{\mathrm{pre}}
  \hat{\mathbf{u}}_{i}^{\mathrm{grv}} .
\end{equation}
Thus, stretching changes the object geometry, scale, and translation, while preserving the corrected rotation from diagnostic simulation.

\paragraph{SAM3D Fine-tuning Details.}
We construct two types of preference pairs for fine-tuning SAM3D under the resampling input in \eqnref{eq:resample}.
First, for occlusion-completion pairs, we synthesize paired-object scenes using a text-to-image generator~\cite{flux-2-2025}, where one object substantially occludes the target object.
An image-to-image editor~\cite{flux-2-2025} then removes the occluding object and completes the full shape of the target object.
From the original occluded image, we extract the modal mask $\vec{M}_i$, while from the edited image, we extract a full-object mask $\vec{M}'_i$ that serves as an amodal crop mask.
We run SAM3D on both images to obtain shape latents: the latent from the occluded input is treated as the rejected sample $x_i^{-}$, and the latent from the edited full-object input is treated as the preferred sample $x_i^{+}$.
During fine-tuning, the condition
$c_i = (\vec{I}, \vec{M}_i, \vec{I}_{\vec{M}'_i}, \vec{D})$
follows the same input setting as \eqnref{eq:resample}.

We additionally construct clean preservation pairs to avoid degrading SAM3D's original input behavior.
For these pairs, we synthesize scenes containing an unoccluded object, obtain its segmentation mask, and run SAM3D to produce a clean latent as $x_i^{+}$.
We then corrupt the mask by dropping part of the segmentation and run SAM3D again, using the resulting latent as $x_i^{-}$.
Thus, occlusion-completion pairs teach amodal completion under the new crop setting, while clean preservation pairs regularize the model to retain its original modal-mask reconstruction ability.
Before training, we filter preferred shape samples using a VLM~\cite{qwen36_35b_a3b} given the original image and six rendered views of each preferred mesh.
We retain only samples that pass checks for a single target, correct identity, complete major parts, and consistency with the visible shape.

We fine-tune SAM3D using the combined objective in \eqnref{eq:total_loss}. Below, we detail its two components:
$\mathcal{L}_{\mathrm{FM\mbox{-}DPO}}$ and $\mathcal{L}_{\mathrm{SFT}}$.
Let $\ell_{\theta}^{\mathrm{shape}}(x,c,t,\epsilon)$ denote the shape-branch flow-matching loss of SAM3D for latent $x$ under condition $c$, timestep $t$, and noise $\epsilon$.
For each preference pair, we compute the win--lose loss difference of the current model as
\begin{equation}\label{eq:supp_model_diff}
  d_{\theta,i}
  =
  \ell_{\theta}^{\mathrm{shape}}(x_i^{+},c_i,t_i,\epsilon_i^{+})
  -
  \ell_{\theta}^{\mathrm{shape}}(x_i^{-},c_i,t_i,\epsilon_i^{-}),
\end{equation}
and the corresponding difference of the frozen reference model as
\begin{equation}\label{eq:supp_ref_diff}
  d_{\mathrm{ref},i}
  =
  \ell_{\theta_{\mathrm{ref}}}^{\mathrm{shape}}(x_i^{+},c_i,t_i,\epsilon_i^{+})
  -
  \ell_{\theta_{\mathrm{ref}}}^{\mathrm{shape}}(x_i^{-},c_i,t_i,\epsilon_i^{-}),
\end{equation}
where $\theta_{\mathrm{ref}}$ is the frozen base SAM3D model.
The current model and the reference model are evaluated on the same condition, timestep, and noise for each preference pair.
The FM-DPO term optimizes preference improvement relative to the frozen reference.
The additional SFT term in \eqnref{eq:total_loss} is defined as
\begin{equation}\label{eq:supp_sft_loss}
  \mathcal{L}_{\mathrm{SFT}}(\theta)
  =
  \mathbb{E}_{i}
  \left[
  w_i\,
  \ell_{\theta}^{\mathrm{shape}}
  (x_i^{+},c_i,t_i,\epsilon_i^{+})
  \right].
\end{equation}
This term directly minimizes the preferred-sample flow-matching loss using the same preferred latent, condition, timestep, and noise as the FM-DPO term.

The sample weight $w_i$ is defined by the pair type:
\begin{equation}\label{eq:supp_pair_weight}
  w_i
  =
  \begin{cases}
  1.0, & \mathrm{if}\ i\ \mathrm{is\ an\ occlusion\mbox{-}completion\ pair}, \\
  0.1, & \mathrm{if}\ i\ \mathrm{is\ a\ clean\ preservation\ pair}.
  \end{cases}
\end{equation}
We apply these pair weights to both the FM-DPO and SFT terms, assigning a larger weight to occlusion-completion pairs because they directly target the amodal resampling failure mode, while clean preservation pairs act as a weaker regularizer.

We do not fine-tune the full SAM3D generator.
Instead, we attach LoRA~\cite{lora} adapters only to the attention projections of the shape generation branch and update these adapters during training.
This limits adaptation to the shape latent pathway while preserving the base model's original reconstruction behavior.

\subsection{Sequential Composition Details}\label{sec:supp_seq}

\noindent\textbf{Ordering strategy.}
Dynamic objects are ordered by the height of their world-frame bounding-box bottom along the up direction, and the lowest object is placed first.

\noindent\textbf{Wall-attached object handling.}
Hanging objects identified by the VLM~\cite{gpt4} (\secref{sec:supp_sim}) remain held by their $K$ anchors during penetration resolution, so a single-pin object retains 3-DoF rotation about its anchor point and a two-pin object retains 1-DoF rotation about the line joining its anchors. Without these anchors, hanging objects released as fully dynamic bodies tend to rotate or fall away from the wall under penetration-resolution forces, since their visible silhouette underdetermines the wall-normal contact geometry.

\noindent\textbf{Final-pass full settling.}
Once sequential penetration resolution has placed every object into a collision-consistent assembled scene, we run a single final settling simulation over the whole scene. Unlike the diagnostic probe, which keeps only one object dynamic and freezes it at first contact, here all \emph{free} objects are simultaneously dynamic, simulated as full 6-DoF rigid bodies and allowed to settle jointly under gravity for a fixed duration $T_\text{settle}$, while \emph{point-anchored} and \emph{line-anchored} objects stay pinned at their anchors and move only by rotation about the anchor point (3-DoF) or anchor line (1-DoF), respectively.
This pass corrects minor residual motion left by sequential placement and yields a globally consistent layout.
Objects whose displacement in this pass exceeds a divergence bound are reverted to their pre-settle pose.

\section{Experimental Details}\label{sec:supp_exp_details}

This section provides details of the quantitative, qualitative, and application experiments.

\subsection{Metrics}
\label{sec:12}

\noindent\textbf{Physical plausibility metrics.}
We evaluate reconstruction quality at \emph{post-sim}, the configuration obtained after a $5\,\mathrm{s}$ MuJoCo gravity-settle simulation, and we report both mask-side and shape-side metrics. For mask-side metrics, we render all meshes jointly and compare their corresponding visible rendered mask against input segmentation. We then report (i) \textbf{2D~IoU}, the mean IoU between each ground-truth mask and its identity-paired prediction, and (ii) \textbf{2D~ABO},
\[
\mathrm{ABO}_{\mathrm{2D}}
\;=\;
\frac{1}{N_{\mathrm{gt}}}\sum_{i=1}^{N_{\mathrm{gt}}}
\max_{j=1,\dots,M_{\mathrm{pred}}} \mathrm{IoU}\!\left(g_i, p_j\right),
\]
with the denominator fixed at the ground-truth count so that missing or filtered-out predictions are penalized as misses rather than hidden.
Shape-side metrics are computed against the ADT object library. Each ground-truth mesh is placed in the scene frame using its provided pose, transformed into the input camera frame, and sampled with $10\mathrm{k}$ surface points. 

We report \textbf{3D~IoU} as the mean axis-aligned-bounding-box (AABB) IoU over matched (GT,~pred) pairs and
\textbf{3D~ABO},
\[
\mathrm{ABO}_{\mathrm{3D}}
\;=\;
\frac{1}{N_{\mathrm{gt}}}\sum_{i=1}^{N_{\mathrm{gt}}}
\max_{j} \mathrm{IoU}_{\mathrm{3D}}\!\left(B_{g_i}, B_{p_j}\right),
\]
again with denominator $N_{\mathrm{gt}}$. We compute all AABBs in the shared input camera frame.

\noindent\textbf{Free/hanging variants (fh/fo).}
We compute each metric under two settings, $\mathrm{fh}$ and $\mathrm{fo}$, and report their average, e.g.\ $\mathrm{ABO}_{\mathrm{fh/fo}}=\tfrac12(\mathrm{ABO}_{\mathrm{fh}}+\mathrm{ABO}_{\mathrm{fo}})$ (likewise for IoU). Both settings keep the denominator $N_{\mathrm{gt}}$ but take the maximum only over predictions that are valid after the evaluation simulation: $\mathrm{fh}$ excludes \emph{penetrating} predictions (defined below), and $\mathrm{fo}$ additionally excludes \emph{hanging} ones, so a ground-truth object without a valid prediction scores zero.

\noindent\textbf{Penetration criterion and Stability.}
An object is \textbf{penetrating} if any of its post-simulation contacts has a penetration depth above $\varepsilon = 2\,\mathrm{mm}$. 
We report \textit{Stab.} as the fraction of all simulated predicted objects that end the evaluation simulation free and stable.
Let $\mathcal{O}$ denote the predicted objects entered into the evaluation simulation, and $\mathcal{O}_{\mathrm{free}} \subseteq \mathcal{O}$ those that are neither hanging nor penetrating at the end of the simulation.
An object is considered stable if its final displacement $\Delta(o)$ from the initial pose is below $5\,\mathrm{cm}$.

\[
\mathrm{Stab.}
=
\frac{
\left|\{o \in \mathcal{O}_{\mathrm{free}} : \Delta(o) < 0.05\,\mathrm{m}\}\right|
}{
\left|\mathcal{O}\right|
}
\times 100.
\]

\subsection{Mesh Quality Evaluation Details}

For each instance that triggers resampling, we render every method's mesh from the input camera and arrange the input image with its visible-mask overlay and two candidate renderings in a horizontal triptych. A VLM judge then selects the more plausible full-object completion of the partially observed occluded instance. Each method pair is queried in both orders to counter left/right position bias.

Given the resulting win counts $A_{ij}$ (number of times method $i$ was preferred over $j$, with ties contributing $+1$ to both $A_{ij}$ and $A_{ji}$), we fit Elo ratings $r_i$, initialized to $1000$, by minimizing the Bradley--Terry negative log-likelihood with the standard Elo scale $c=400$,
\begin{equation}
\mathcal{L}(r) \;=\;
\sum_{i \neq j} A_{ij}\,
\log\!\bigl(1 + 10^{(r_j - r_i)/c}\bigr),
\quad
P(i \succ j) \;=\; \frac{1}{1 + 10^{(r_j - r_i)/c}}.
\label{eq:elo}
\end{equation}

\subsection{Aria Digital Twin Preprocessing Details} \label{sec:11}
We build a scenario-balanced evaluation set of $40$ frames drawn from $24$ Aria Digital Twin (ADT)~\cite{pan2023ariadigitaltwin} sequences, selected with five sequences each from the \textit{clean}, \textit{decoration}, \textit{meal}, and \textit{work} scenarios and four from \textit{recognition}. 
We exclude the multi-user sequences, where third-party hands appear in frame, and the skeleton-annotated releases, whose motion-capture skeleton pixels contaminate the ground-truth segmentation. 
Within each retained sequence, we pick up to two frames maximally separated in time that satisfy five conditions. (i)~GT depth, segmentation, and 3D bounding boxes are available. (ii)~The head-mounted camera is near-stationary, with linear and angular velocities below $0.6\,\mathrm{m/s}$ and $0.6\,\mathrm{rad/s}$. (iii)~Every movable object is at rest, moving less than $10\,\mathrm{mm}$ within $\pm 5\,\mathrm{s}$, so that the GT 3D state matches the RGB frame. (iv)~At least ten non-structural objects are visible, with at least $60\%$ of their pixels inside the central $60\%{\times}60\%$ region. (v)~MediaPipe~\cite{mediapipe} detects no hands in the frame or its $90^{\circ}$-rotated version. 
Each accepted frame is then undistorted to a pinhole camera, cropped to the largest valid square, and rotated $90^{\circ}$ clockwise so that gravity points downward.

Ground-truth per-object masks are taken from the ADT instance segmentation, excluding architectural elements (floors, walls, doors, and ceilings), instances without a GT 3D bounding box, and instances smaller than $100$ pixels or $0.1\%$ of the image. Built-in furniture and large appliances (e.g., beds, kitchen islands, and refrigerators) are kept as physical objects, yielding on average $26$ objects per scene, which are passed to SAM3D to generate the input meshes.

\subsection{Application Details}

\noindent\textbf{Robot-arm Manipulation.}
We adopt the closed-loop VLM agent of VLMPose~\cite{vlmpose} as our text-guided 6D goal-pose predictor, using GPT-5.2 with the authors' default configuration. The predicted goal pose is passed to GraspNet~\cite{fang2020graspnet} for grasp proposals on the target object's mesh-sampled point cloud, and OMPL~\cite{sucan2012open} plans a rectangular trajectory that lifts, translates, and lowers the object with a Franka arm to reduce inter-object collisions.

\section{Additional Experimental Results}
\subsection{Additional Quantitative Comparisons}
We benchmark Ours against SAM3D~\cite{sam3d}, Gen3DSR~\cite{ardelean2025gen3dsr}, 3D-RE-GEN~\cite{sautter20253dregen}, and 3D-Fixer~\cite{3D-Fixer} on GraspClutter6D~\cite{back2025graspclutter6d}, Aria Digital Twin (ADT)~\cite{pan2023ariadigitaltwin}, and \dataname, reporting 2D IoU (fh/fo and identity-paired) and Stab., plus 3D IoU on ADT.
For Gen3DSR and 3D-RE-GEN we also evaluate a variant that swaps in our background reconstruction (``our bg'') to isolate the foreground predictor from the layout.

\tabref{tab:supp_quant_iou} shows that Ours achieves the highest IoU$_{\mathrm{fh/fo}}$ and Stab.\ on all three datasets and the highest 3D IoU$_{\mathrm{fh/fo}}$ and 3D IoU on ADT, with Stab.\ margins of 30.3 to 51.7 percentage points over the strongest baseline.
On identity-paired IoU, Ours leads on GraspClutter6D and \dataname{}, while Gen3DSR scores higher on ADT (0.548 vs.\ 0.275) because most of its objects interpenetrate (93.3\% in \tabref{tab:main_quant}) and stay fused near their initial poses, which preserves their projected masks.
The Stab.\ margin reflects the role of gravity-based diagnostics and shape correction in producing layouts that physically settle, rather than merely projecting plausibly into the input view (see also \tabref{tab:ablation} and \tabref{tab:supp_abl_real}).

\subsection{Correction Threshold Analysis}\label{sec:supp_tau}
We compare gravity-axis stretching and amodal shape resampling on 708 ADT objects to examine the correction threshold $\tau = 0.15$.
The objects are partitioned into four disjoint ranges of the normalized gravity-axis displacement $\rho_i$ defined in \eqnref{eq:stretch_ratio}.
Since ADT provides ground-truth (GT) object meshes, we evaluate shape quality directly by aligning each reconstructed mesh to its GT mesh with ICP and normalizing ADD-S by the GT object diameter.
As shown in \tabref{tab:supp_tau}, the two branches yield nearly identical error for $\rho_i < 0.15$ (0.0277 vs.\ 0.0281), whereas resampling reduces the error by 15.3\%, 13.3\%, and 62.8\% in the three ranges with $\rho_i \geq 0.15$.
These results support stretching below $\tau = 0.15$ and resampling at or above it on this evaluation set.
The resulting trade-off between shape quality and resampling time is summarized in \tabref{tab:tau_main}.

\begin{table}[!ht]
  \centering
  \setlength{\abovecaptionskip}{7pt}%
  \small
  \begin{tabular}{lcccc}
    \toprule
    $\rho_i$ range & \#Objects & \multicolumn{2}{c}{ADD-S$_{\mathrm{norm}}$ $\downarrow$} & Change \\
    \cmidrule(lr){3-4}
    & & Stretch & Resample & \\
    \midrule
    $\rho_i < 0.15$           & 226 & \textbf{0.0277} & 0.0281 & $+1.4\%$ \\
    $0.15 \leq \rho_i < 0.30$ & 108 & 0.0321 & \textbf{0.0272} & $-15.3\%$ \\
    $0.30 \leq \rho_i < 1.0$  & 120 & 0.0353 & \textbf{0.0306} & $-13.3\%$ \\
    $\rho_i \geq 1.0$         & 254 & 0.1131 & \textbf{0.0421} & $-62.8\%$ \\
    \midrule
    All evaluated objects     & 708 & 0.0603 & \textbf{0.0334} & $-44.6\%$ \\
    \bottomrule
  \end{tabular}
  \caption{\textbf{Shape quality of stretching and resampling across $\rho_i$ ranges on ADT.} ADD-S$_{\mathrm{norm}}$ is ADD-S after rigid ICP alignment to the GT mesh, normalized by the GT object diameter. Change is the relative error of resampling with respect to stretching, so negative values indicate lower error with resampling. Bold marks the lower error in each row.}
  \label{tab:supp_tau}
\end{table}

\subsection{Progressive Ablation on Real Data}\label{sec:supp_abl_real}
We repeat the progressive ablation of \tabref{tab:ablation} on GraspClutter6D and ADT, adding one stage at a time to the SAM3D baseline.
As shown in \tabref{tab:supp_abl_real}, alignment and penetration resolution each increase ABO and ABO$_{\mathrm{fh/fo}}$ on both datasets, and gravity correction reduces $D_{\mathrm{mean}}$ from 0.342 to 0.234 on GraspClutter6D and from 0.586 to 0.357 on ADT.
As on \dataname{}, the full model achieves the highest ABO and ABO$_{\mathrm{fh/fo}}$ on both datasets, and it also attains the lowest $D_{\mathrm{mean}}$ on ADT.
Two trade-offs appear. On ADT, gravity correction slightly lowers ABO from 0.216 to 0.211, and on GraspClutter6D, shape correction raises $D_{\mathrm{mean}}$ from 0.234 to 0.250.

\begin{table}[h]
  \centering
  \setlength{\abovecaptionskip}{7pt}%
  \small
  \begin{tabular}{l|ccc|ccc}
    \toprule
    & \multicolumn{3}{c|}{GraspClutter6D} & \multicolumn{3}{c}{ADT} \\
    Variants
    & ABO $\uparrow$
    & ABO$_{\mathrm{fh/fo}}$ $\uparrow$
    & $D_{\mathrm{mean}}$ $\downarrow$
    & ABO $\uparrow$
    & ABO$_{\mathrm{fh/fo}}$ $\uparrow$
    & $D_{\mathrm{mean}}$ $\downarrow$ \\
    \midrule
    SAM3D       & 0.121 & 0.087 & 0.638 & 0.147 & 0.092 & 0.721 \\
    \ +Align.   & 0.186 & 0.150 & 0.494 & 0.189 & 0.138 & 0.630 \\
    \ +Pen.     & \cellcolor{tabthird}0.240 & \cellcolor{tabthird}0.223 & \cellcolor{tabthird}0.342 & \cellcolor{tabsecond}0.216 & \cellcolor{tabsecond}0.165 & \cellcolor{tabthird}0.586 \\
    \ +Grv.     & \cellcolor{tabsecond}0.258 & \cellcolor{tabsecond}0.242 & \cellcolor{tabfirst}\textbf{0.234} & \cellcolor{tabthird}0.211 & \cellcolor{tabsecond}0.165 & \cellcolor{tabsecond}0.357 \\
    \midrule
    \ +Re. (Full) & \cellcolor{tabfirst}\textbf{0.267} & \cellcolor{tabfirst}\textbf{0.248} & \cellcolor{tabsecond}0.250 & \cellcolor{tabfirst}\textbf{0.299} & \cellcolor{tabfirst}\textbf{0.256} & \cellcolor{tabfirst}\textbf{0.311} \\
    \bottomrule
  \end{tabular}
  \caption{\textbf{Progressive ablation on real data.} Each row adds one stage to the row above, following \tabref{tab:ablation}. Align., Pen., Grv., and Re.\ denote alignment, penetration resolution, gravity correction, and shape correction with resampling.}
  \label{tab:supp_abl_real}
\end{table}

\subsection{Upstream Module Ablations}\label{sec:supp_upstream}
We examine the sensitivity of our pipeline to two upstream modules on \dataname{} while keeping the downstream pipeline unchanged.
First, we replace FoundationPose~\cite{wen2024foundationpose} with GoTrack~\cite{nguyen2025gotrackgeneric6dofobject} for object pose initialization.
Second, we reconstruct the base structures from the original image instead of the image decluttered with Gemini~\cite{gemini}.
As shown in \tabref{tab:supp_upstream}, replacing FoundationPose with GoTrack lowers ABO$_{\mathrm{fh/fo}}$ from 0.507 to 0.260 and raises $D_{\mathrm{mean}}$ from 0.128 to 0.337, $E_{\mathrm{pk}}$ from 0.42 to 1.65, and Penetr.\ from 2.5\% to 5.6\%.
Removing decluttering lowers ABO$_{\mathrm{fh/fo}}$ from 0.507 to 0.444 and raises $D_{\mathrm{mean}}$ from 0.128 to 0.181.
Of the two modules, pose initialization has the larger impact in these experiments, which makes it the more critical upstream bottleneck.
These ablations are not an exhaustive oracle replacement of every pretrained module, but they show how changes in these two components affect the final reconstruction.

\begin{table}[h]
  \centering
  \setlength{\abovecaptionskip}{7pt}%
  \small
  \begin{tabular}{lccccc}
    \toprule
    Variant & ABO$_{\mathrm{fh/fo}}$ $\uparrow$ & ABO $\uparrow$ & $D_{\mathrm{mean}}$ $\downarrow$ & $E_{\mathrm{pk}}$ $\downarrow$ & Penetr.\ $\downarrow$ \\
    \midrule
    \multicolumn{6}{l}{\textit{Pose initialization}} \\
    FoundationPose (Ours) & \textbf{0.507} & \textbf{0.532} & \textbf{0.128} & \textbf{0.42} & \textbf{2.5\%} \\
    GoTrack               & 0.260 & 0.290 & 0.337 & 1.65 & 5.6\% \\
    \midrule
    \multicolumn{6}{l}{\textit{Base decluttering}} \\
    w/ Gemini (Ours)      & \textbf{0.507} & \textbf{0.532} & \textbf{0.128} & -- & -- \\
    w/o Gemini            & 0.444 & 0.473 & 0.181 & -- & -- \\
    \bottomrule
  \end{tabular}
  \caption{\textbf{Upstream module ablations on \dataname.} Each block changes one upstream module and keeps the rest of the pipeline fixed. Bold marks the best value in each column within each block. Dashes denote metrics not reported for the decluttering ablation.}
  \label{tab:supp_upstream}
\end{table}

\subsection{Robustness to Pose Perturbation}\label{sec:supp_pose_perturb}
To evaluate robustness to inaccurate pose initialization, we perturb the initial rotation of each object on \dataname{} before the subsequent correction stages.
The perturbation rotates the object about a randomly sampled axis by an angle drawn from a Gaussian distribution with mean $45^{\circ}$ and standard deviation $15^{\circ}$.
As shown in \tabref{tab:supp_pose_perturb}, stretching alone improves all three metrics over no shape correction, which corresponds to the +Grv.\ setting of \tabref{tab:ablation}.
Adding resampling improves them further, raising ABO$_{\mathrm{fh/fo}}$ from 0.3107 to 0.3320 and lowering $D_{\mathrm{mean}}$ from 0.4771 to 0.3905.
All perturbed variants still have lower ABO and ABO$_{\mathrm{fh/fo}}$ and higher $D_{\mathrm{mean}}$ than the unperturbed full model in \tabref{tab:ablation} (0.532, 0.507, and 0.128).
Shape correction therefore partially mitigates inaccurate pose initialization rather than fully recovering from it.

\begin{table}[h]
  \centering
  \setlength{\abovecaptionskip}{7pt}%
  \small
  \begin{tabular}{lccc}
    \toprule
    Correction & ABO$_{\mathrm{fh/fo}}$ $\uparrow$ & ABO $\uparrow$ & $D_{\mathrm{mean}}$ $\downarrow$ \\
    \midrule
    No shape correction        & 0.2512 & 0.2699 & 0.5318 \\
    Stretch only               & 0.3107 & 0.3304 & 0.4771 \\
    Stretch + Resample (Ours)  & \textbf{0.3320} & \textbf{0.3576} & \textbf{0.3905} \\
    \bottomrule
  \end{tabular}
  \caption{\textbf{Recovery under rotation perturbation on \dataname.} The initial rotation of each object is perturbed about a random axis by an angle drawn from a Gaussian distribution with mean $45^{\circ}$ and standard deviation $15^{\circ}$. Bold marks the best value in each column.}
  \label{tab:supp_pose_perturb}
\end{table}

\subsection{Appearance Metrics}\label{sec:supp_appearance}
ABO measures mask overlap with the input view and therefore does not depend on the textures of the generated objects, which are not the focus of our work.
For completeness, we also report the appearance metrics PSNR, LPIPS, and CLIP distance on GraspClutter6D and \dataname{}, which are strongly affected by these generated textures.
As shown in \tabref{tab:supp_appearance}, Ours achieves the best value on all three metrics on both datasets.

\begin{table}[h]
  \centering
  \setlength{\abovecaptionskip}{7pt}%
  \small
  \resizebox{\linewidth}{!}{%
  \begin{tabular}{lccccccc}
    \toprule
     & \multicolumn{3}{c}{GraspClutter6D} & & \multicolumn{3}{c}{\dataname} \\
    \cmidrule(lr){2-4} \cmidrule(lr){6-8}
    Method & PSNR $\uparrow$ & LPIPS $\downarrow$ & CLIP $\downarrow$ & & PSNR $\uparrow$ & LPIPS $\downarrow$ & CLIP $\downarrow$ \\
    \midrule
    SAM3D~\cite{sam3d} & \cellcolor{tabsecond}11.09 & 0.381 & \cellcolor{tabsecond}0.272 & & \cellcolor{tabsecond}11.59 & \cellcolor{tabsecond}0.405 & \cellcolor{tabsecond}0.220 \\
    Gen3DSR~\cite{ardelean2025gen3dsr} (our bg) & 10.97 & 0.387 & 0.357 & & 10.34 & 0.449 & 0.344 \\
    Gen3DSR~\cite{ardelean2025gen3dsr} & 10.91 & 0.379 & 0.357 & & 9.88 & 0.444 & 0.395 \\
    3D-RE-GEN~\cite{sautter20253dregen} (our bg) & 10.98 & 0.381 & 0.358 & & 10.73 & 0.426 & 0.357 \\
    3D-RE-GEN~\cite{sautter20253dregen} & \cellcolor{tabthird}11.08 & \cellcolor{tabthird}0.378 & 0.371 & & 9.88 & 0.438 & 0.423 \\
    3D-Fixer~\cite{3D-Fixer} & 10.97 & \cellcolor{tabsecond}0.367 & \cellcolor{tabthird}0.294 & & \cellcolor{tabthird}10.85 & \cellcolor{tabthird}0.411 & \cellcolor{tabthird}0.288 \\
    \midrule
    \textbf{Ours} & \cellcolor{tabfirst}\textbf{13.65} & \cellcolor{tabfirst}\textbf{0.306} & \cellcolor{tabfirst}\textbf{0.207} & & \cellcolor{tabfirst}\textbf{14.41} & \cellcolor{tabfirst}\textbf{0.314} & \cellcolor{tabfirst}\textbf{0.138} \\
    \bottomrule
  \end{tabular}%
  }
  \caption{\textbf{Appearance metrics on GraspClutter6D and \dataname.} CLIP denotes CLIP distance. Columns marked $\uparrow$ ($\downarrow$) indicate higher (lower) is better.}
  \label{tab:supp_appearance}
\end{table}

\subsection{Runtime Analysis}\label{sec:supp_runtime}
We measure the per-stage runtime of our pipeline on \dataname{} and report it in \tabref{tab:supp_runtime}.
GPU-based stages run on a single NVIDIA RTX PRO 6000 Blackwell GPU, the same GPU used for training, and the MLLM and VLM stages report end-to-end external API latency.
The full pipeline takes 347.7\,s per scene, about 2.13$\times$ the 163.5\,s of preprocessing plus SAM3D lifting.
The largest additional costs come from resampling (52.4\,s), decluttered image generation (46.9\,s), and FoundationPose pose alignment (38.9\,s), while plane extraction and V-HACD decomposition add little overhead.

\begin{table}[h]
  \centering
  \setlength{\abovecaptionskip}{7pt}%
  \small
  \begin{tabular}{llcc}
    \toprule
    \# & Stage & Time (s) & Relative \\
    \midrule
    1 & Preprocessing + SAM3D lifting (baseline) & 163.5 $\pm$ 27.3 & 100\% \\
      & \quad Preprocessing (depth + masks) & 4.7 $\pm$ 7.6 & 2.9\% \\
      & \quad SAM3D lifting (incl.\ base) & 158.8 $\pm$ 26.2 & 97.1\% \\
    2 & Decluttered image generation (MLLM) & 46.9 $\pm$ 7.0 & 28.7\% \\
    3 & Pose-DoF estimation (VLM) & 18.1 $\pm$ 10.4 & 11.1\% \\
    4 & Plane extraction & 2.4 $\pm$ 1.9 & 1.5\% \\
    5 & V-HACD decomposition & 5.7 $\pm$ 3.3 & 3.5\% \\
    6 & FoundationPose pose alignment & 38.9 $\pm$ 4.6 & 23.8\% \\
    7 & Physics simulation + setup & 19.8 $\pm$ 3.6 & 12.1\% \\
    8 & Resampling & 52.4 $\pm$ 13.6 & 32.0\% \\
    \midrule
      & Total & 347.7 $\pm$ 45 & 212.7\% \\
    \bottomrule
  \end{tabular}
  \caption{\textbf{Per-stage runtime on \dataname.} Time is mean $\pm$ std per scene. Relative runtime is normalized to preprocessing plus SAM3D lifting. MLLM and VLM stages report external API latency.}
  \label{tab:supp_runtime}
\end{table}

\subsection{Additional Qualitative Comparisons}
We supplement the qualitative comparisons in the main paper with further results on the Aria Digital Twin (ADT)~\cite{pan2023ariadigitaltwin} and GraspClutter6D~\cite{back2025graspclutter6d} datasets.
\figref{fig:supp_qual_adt} extends the ADT comparisons, while \figref{fig:supp_qual_decluttered6d} extends those on GraspClutter6D.

\input{figures/110_supp10/figure}
\vspace{-12mm}
\input{figures/111_supp11/figure}

%% file: figures/100_supp0/figure.tex
\begin{figure}[t!]
  \centering
  \includegraphics[width=\textwidth]{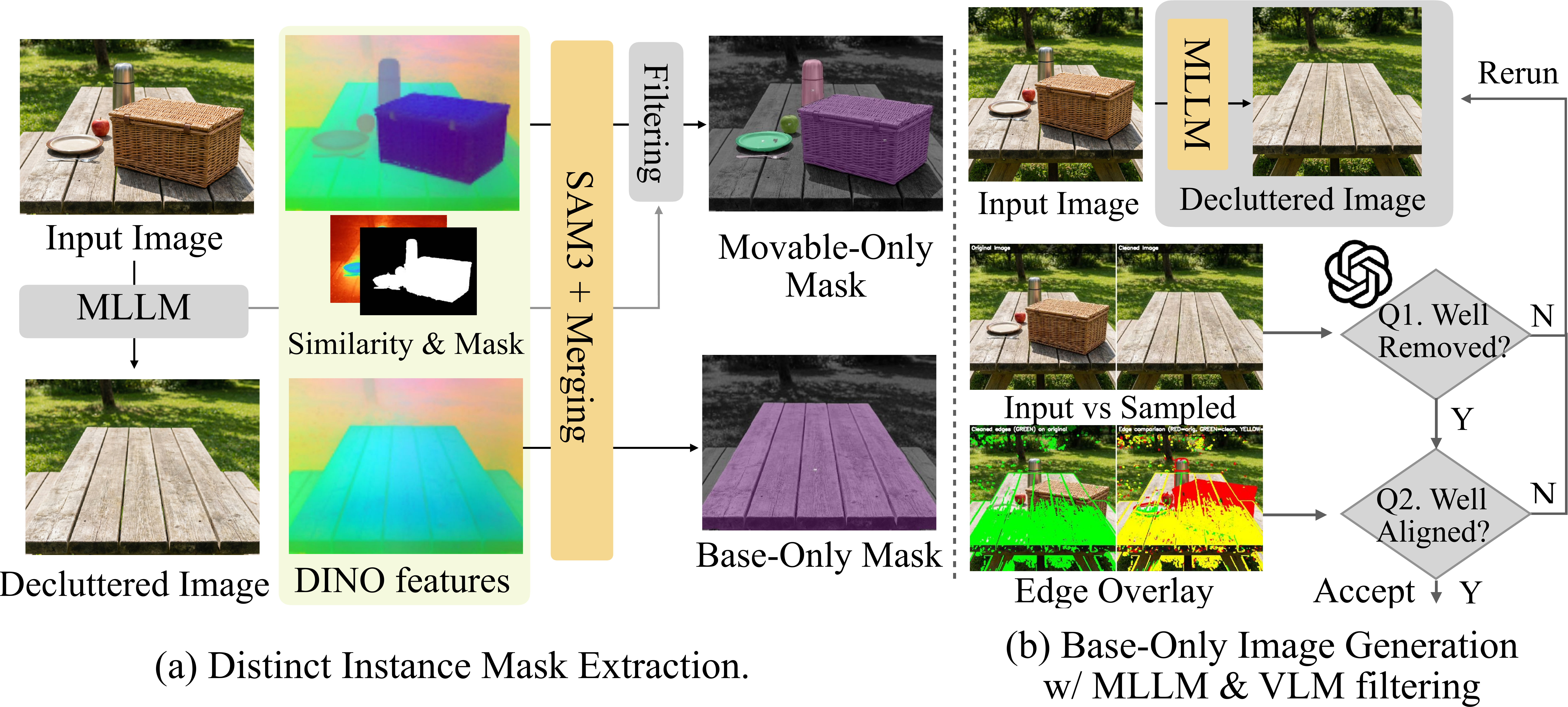}
  \caption{\textbf{Distinct Instance Extraction and Decluttered Image $I_{\text{base}}$ Generation.}
(a) We generate a decluttered image $I_{\text{base}}$ via MLLM~\cite{gemini} and compute an object probability map from pixel differences with the original image. Candidate masks in the original image are filtered by their overlap with this map to identify liftable instances, while masks in the decluttered image are selected using Set-of-Mark-based VLM filtering~\cite{som_prompt} due to the simplified scene layout.
(b) We generate a decluttered image by using an MLLM to remove foreground objects. A VLM gate verifies removal completeness and edge alignment with the input, and the generation–verification process is repeated until the acceptance criteria are met.}
\label{fig:instance_extraction}
\end{figure}

%% file: tables/100_supp_quant1.tex
\begin{table*}[t]
  \centering
  \resizebox{\textwidth}{!}{%
  \begin{tabular}{lccccccccccccc}
    \toprule
     & \multicolumn{3}{c}{GraspClutter6D} & & \multicolumn{5}{c}{AriaDigitalTwin} & & \multicolumn{3}{c}{GenWild} \\
    \cmidrule(lr){2-4} \cmidrule(lr){6-10} \cmidrule(lr){12-14}
    Method
      & IoU$_{\mathrm{fh/fo}}$ $\uparrow$
      & IoU $\uparrow$
      & Stab.\ $\uparrow$
      & 
      & IoU$_{\mathrm{fh/fo}}$ $\uparrow$
      & IoU $\uparrow$
      & 3D IoU$_{\mathrm{fh/fo}}$ $\uparrow$
      & 3D IoU $\uparrow$
      & Stab.\ $\uparrow$
      & 
      & IoU$_{\mathrm{fh/fo}}$ $\uparrow$
      & IoU $\uparrow$
      & Stab.\ $\uparrow$ \\
    \midrule
    SAM3D~\cite{sam3d} & \cellcolor{tabthird}0.055 & 0.081 & 0.7\% &  & \cellcolor{tabthird}0.064 & 0.115 & 0.027 & 0.040 & \cellcolor{tabthird}0.5\% &  & \cellcolor{tabsecond}0.205 & \cellcolor{tabthird}0.246 & 3.2\% \\
    Gen3DSR~\cite{ardelean2025gen3dsr} (our bg) & 0.007 & 0.020 & 0.0\% &  & 0.062 & 0.109 & \cellcolor{tabthird}0.064 & 0.088 & 0.0\% &  & 0.086 & 0.118 & 0.4\% \\
    Gen3DSR~\cite{ardelean2025gen3dsr} & 0.005 & \cellcolor{tabsecond}0.197 & 0.0\% &  & 0.003 & \cellcolor{tabfirst}\textbf{0.548} & 0.001 & \cellcolor{tabsecond}0.220 & 0.0\% &  & 0.036 & \cellcolor{tabsecond}0.458 & 0.0\% \\
    3D-RE-GEN~\cite{sautter20253dregen} (our bg) & 0.016 & 0.032 & 1.1\% &  & 0.041 & 0.061 & 0.036 & 0.051 & 0.0\% &  & 0.090 & 0.130 & \cellcolor{tabthird}5.6\% \\
    3D-RE-GEN~\cite{sautter20253dregen} & 0.028 & 0.054 & \cellcolor{tabsecond}4.4\% &  & 0.025 & 0.065 & 0.019 & 0.042 & 0.1\% &  & 0.081 & 0.135 & \cellcolor{tabsecond}6.7\% \\
    3D-Fixer~\cite{3D-Fixer} & \cellcolor{tabsecond}0.088 & \cellcolor{tabthird}0.100 & \cellcolor{tabthird}3.2\% &  & \cellcolor{tabsecond}0.090 & \cellcolor{tabthird}0.132 & \cellcolor{tabsecond}0.110 & \cellcolor{tabthird}0.153 & \cellcolor{tabsecond}0.7\% &  & \cellcolor{tabthird}0.173 & 0.237 & 2.4\% \\
    \midrule
    \textbf{Ours} & \cellcolor{tabfirst}\textbf{0.213} & \cellcolor{tabfirst}\textbf{0.227} & \cellcolor{tabfirst}\textbf{34.7\%} &  & \cellcolor{tabfirst}\textbf{0.234} & \cellcolor{tabsecond}0.275 & \cellcolor{tabfirst}\textbf{0.239} & \cellcolor{tabfirst}\textbf{0.277} & \cellcolor{tabfirst}\textbf{31.5\%} &  & \cellcolor{tabfirst}\textbf{0.494} & \cellcolor{tabfirst}\textbf{0.519} & \cellcolor{tabfirst}\textbf{58.4\%} \\
    \bottomrule
  \end{tabular}%
  }
  \caption{\textbf{Supplementary IoU-based quantitative comparison.} 2D IoU on GraspClutter6D~\cite{back2025graspclutter6d} and \dataname{}, and both 2D IoU and 3D IoU on AriaDigitalTwin~\cite{pan2023ariadigitaltwin}. All IoUs are measured after the evaluation simulation, and ground-truth objects without a prediction count as zero overlap. Stab.\ is the ratio of predicted objects that remain free and stable after the evaluation simulation.}
  \label{tab:supp_quant_iou}
\end{table*}

%% file: figures/102_wall_hang_supp/figure.tex
\begin{figure}[t!]
  \centering
  \includegraphics[width=\textwidth]{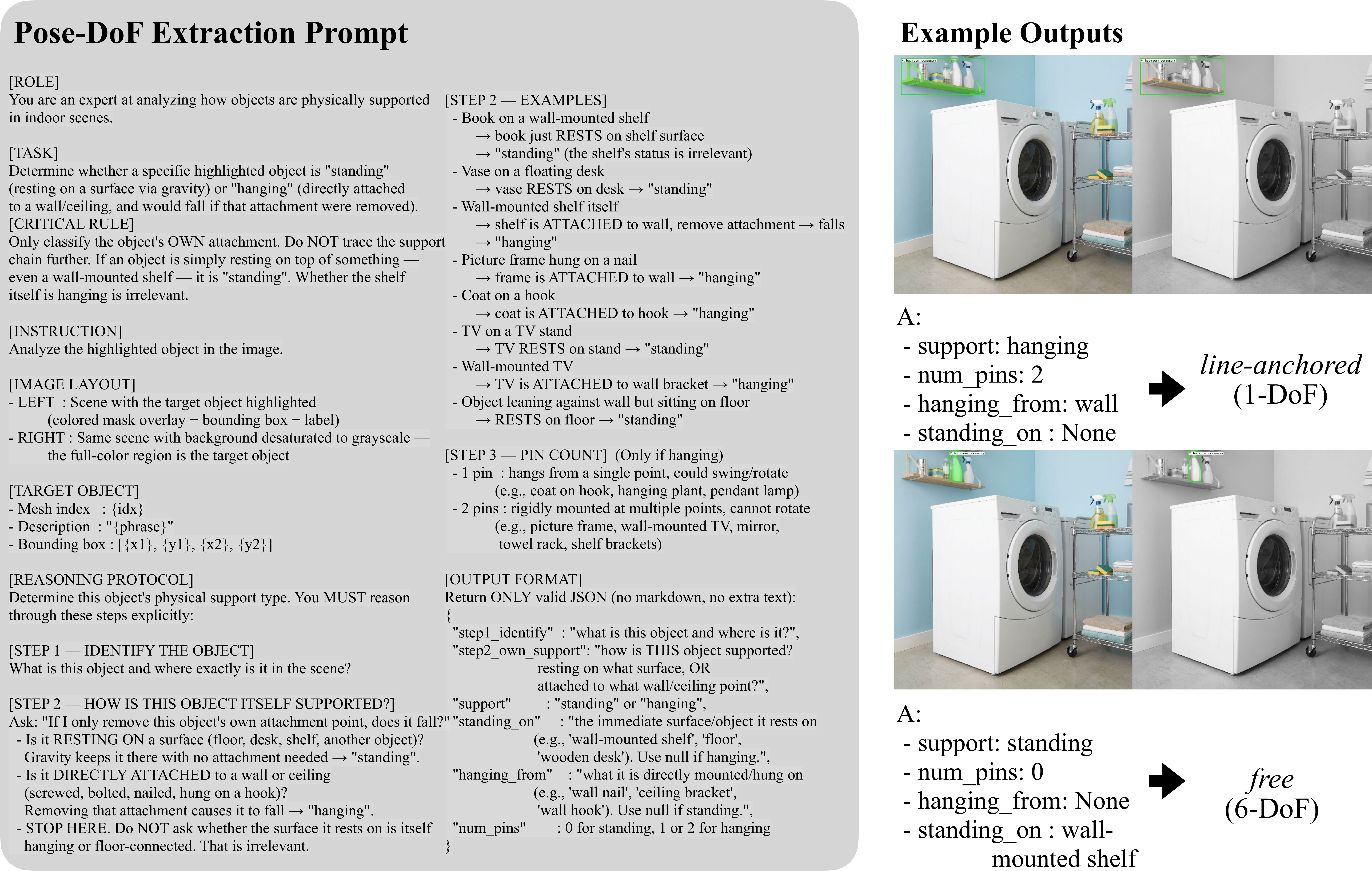}
  \caption{\textbf{VLM-based wall-hang detection.}
Each object is rendered into a two-panel query image with a highlighted overlay (left) and a color-on-grayscale isolation (right), and a VLM classifies it as standing or hanging under an own-attachment-only rule. Hanging objects are additionally tagged with a pin count (0: free, 1: point-anchored, 2: line-anchored).}
\label{fig:wall_hang_supp}
\end{figure}

%% file: figures/110_supp10/figure.tex
\begin{figure}[t!]
  \centering
  \includegraphics[width=\textwidth]{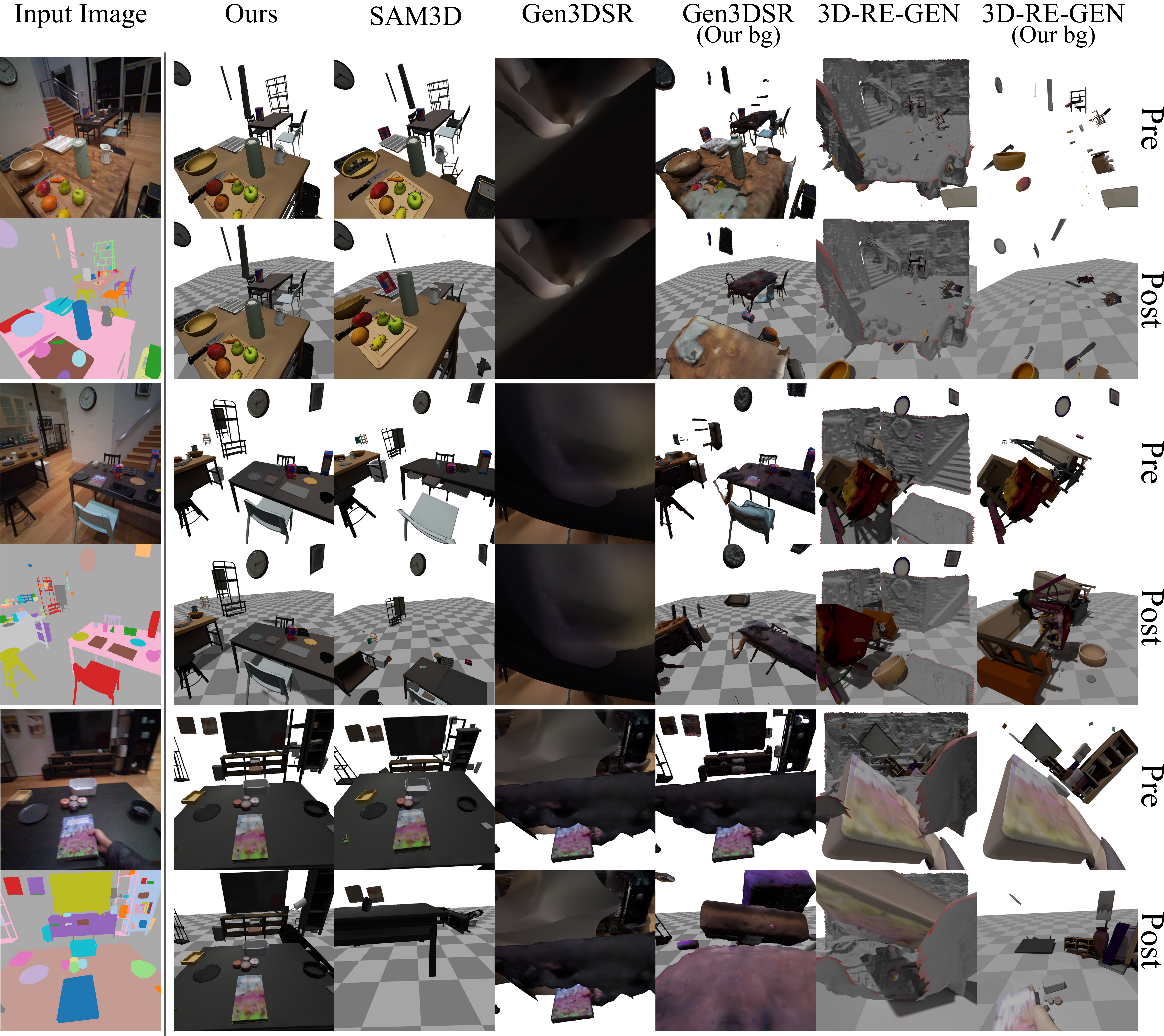}
  \caption{\textbf{Qualitative Comparison of ADT.}
Additional qualitative comparisons on the ADT dataset.}
\label{fig:supp_qual_adt}
\end{figure}

%% file: figures/111_supp11/figure.tex
\begin{figure}[t!]
  \centering
  \includegraphics[width=\textwidth]{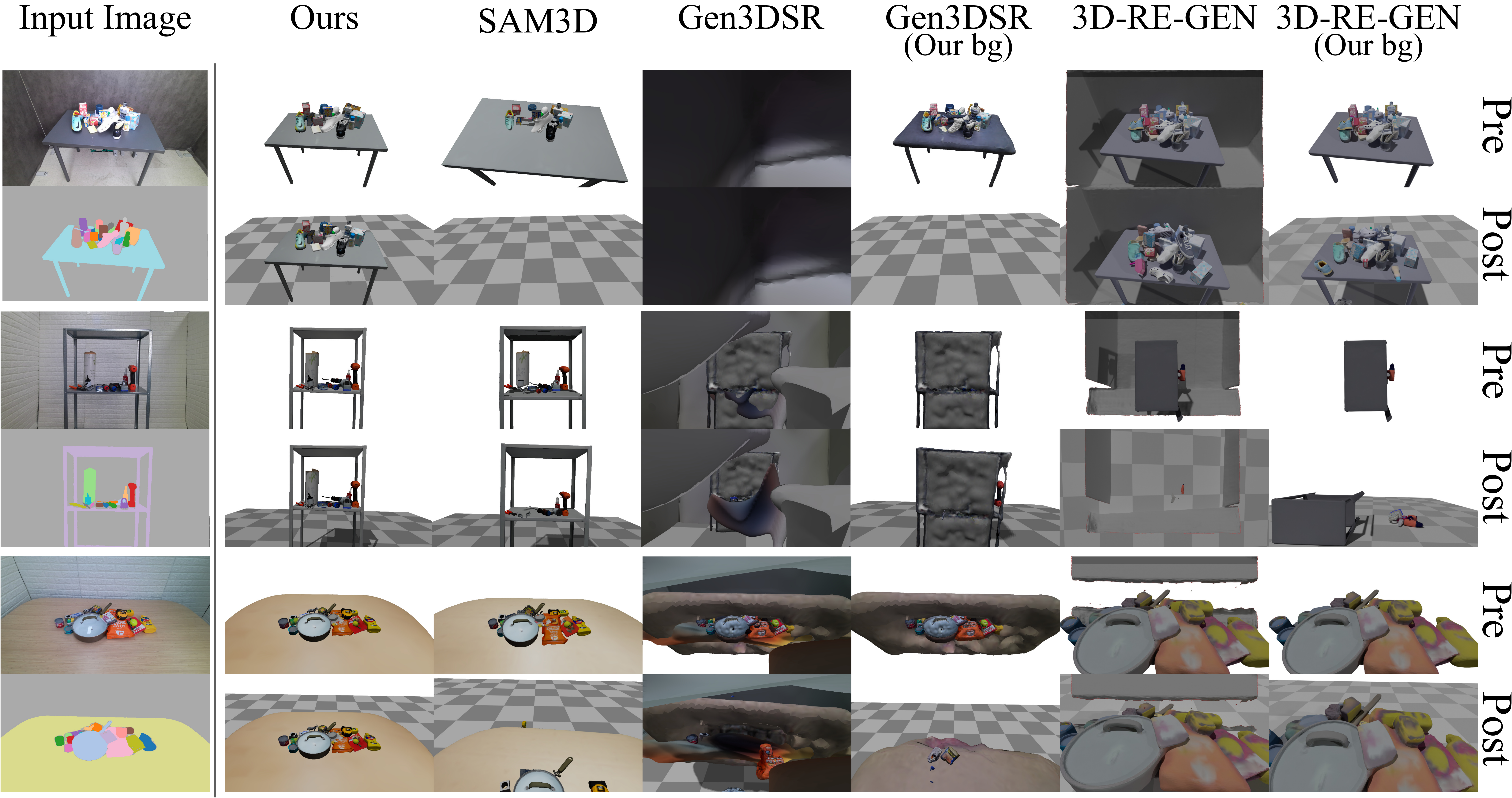}
  \caption{\textbf{Qualitative Comparison of GraspClutter6D.}
Additional qualitative comparisons on the GraspClutter6D dataset.}
\label{fig:supp_qual_decluttered6d}
\end{figure}